\documentclass{article}

\usepackage[preprint]{neurips_2026}

\usepackage[utf8]{inputenc} 
\usepackage[T1]{fontenc}    
\usepackage{hyperref}       
\usepackage{url}            
\usepackage{booktabs}       
\usepackage{amsfonts}       
\usepackage{nicefrac}       
\usepackage{microtype}      
\usepackage{xcolor}         
\usepackage{subcaption}
\usepackage{graphicx}
\usepackage{float}
\usepackage{amssymb}
\usepackage{amsmath}
\usepackage{bbm}
\usepackage{enumitem}
\usepackage{booktabs}
\usepackage{multirow}
\usepackage[table]{xcolor}
\usepackage{algorithm}
\usepackage{algpseudocode}
\usepackage{titletoc}
\usepackage{amsmath}
\usepackage{amssymb}
\usepackage{amsfonts}
\usepackage{wrapfig}

\definecolor{cmlabPrefix}{HTML}{0D207F}
\definecolor{cmlabNumber}{HTML}{0D207F}

\hypersetup{
  colorlinks=true,
  linkcolor=cmlabNumber,
  citecolor=cmlabNumber,
  urlcolor=cmlabNumber,
  pdfborder={0 0 0},
}

\usepackage[capitalize,noabbrev]{cleveref}

\newcommand{\cmlabRef}[4]{%
  {\color{cmlabPrefix}#1}~#3{\color{cmlabNumber}#2}#4%
}
\newcommand{\cmlabRefParen}[4]{%
  {\color{cmlabPrefix}#1}~#3{\color{cmlabNumber}(#2)}#4%
}

\definecolor{lightpeach}{RGB}{255, 218, 185}  
\definecolor{customblue}{RGB}{197, 224, 248}  
\definecolor{custompink}{RGB}{255, 204, 213}  
\definecolor{sh_blue}{RGB}{30, 100, 200}      
\definecolor{sh_red}{RGB}{200, 40, 60}        

\title{DiTTo: Scalable Order-aware All-in-One Image Restoration Agent}

\author{%
  David S.~Hippocampus\thanks{Use footnote for providing further information
    about author (webpage, alternative address)---\emph{not} for acknowledging
    funding agencies.} \\
  Department of Computer Science\\
  Cranberry-Lemon University\\
  Pittsburgh, PA 15213 \\
  \texttt{hippo@cs.cranberry-lemon.edu} \\
}

\cmlabAuthors{Seungho Choi \qquad Jihyong Oh$^{\dagger}$}
\cmlabAffiliations{CMLab, Chung-Ang University}
\cmlabAuthorEmail{\{choiseungho1019,jihyongoh\}@cau.ac.kr}
\cmlabProjectPage{https://cmlab-korea.github.io/DiTTo/}

\begin{document}

\maketitle

\begingroup
\renewcommand{\thefootnote}{}
\footnotetext{$^{\dagger}$ Corresponding author}
\endgroup

\begin{abstract}
Real-world images rarely suffer from a single degradation, and the order in which degradations are removed substantially affects the final restoration quality, motivating agent-based image restoration (IR), where a vision-language model schedules a pool of pre-built restoration-experts.
However, existing training-based agents require $\mathcal{O}((N^{\mathbf{D}})^{2})$ restoration-expert calls per image to construct the Optimal Restoration-action Trajectory Dataset (ORTD), where $N^{\mathbf{D}}$ denotes the number of degradation types in the universe $\mathbf{D}$, and couple agent training to a fixed restoration-expert pool, preventing extension to newly introduced restoration-experts without full retraining.
To overcome these efficiency and extensibility bottlenecks, we propose \textbf{DiTTo}, a novel order-aware image restoration agent framework consisting of the DiTTo Simulator and the DiTTo Agent.
The DiTTo Simulator combines $\cup$S-IR for single-step restoration-action simulation and AiO-IQA for per-action quality prediction, reducing ORTD construction to $\mathcal{O}(N^{\mathbf{D}})$ simulator calls per image; the DiTTo Agent is trained by SFT on the simulator-generated ORTD, followed by \textbf{Order-aware Restoration Alignment (ORA)} that aligns degradation identification, restoration-action-ordering, and output format along independent axes.
This enables \textbf{plug-and-play scalable extensibility}: adding a new restoration-expert requires updating only the lightweight ORA stage.
On the MiO-100 evaluation set with up to five concurrent degradations, our DiTTo Agent achieves state-of-the-art multi-degradation restoration quality among previous agent-based IR methods.
\end{abstract}

\begin{figure}[h]
    \centering
    \begin{minipage}{0.39\linewidth}
        \centering
        \includegraphics[width=\linewidth]{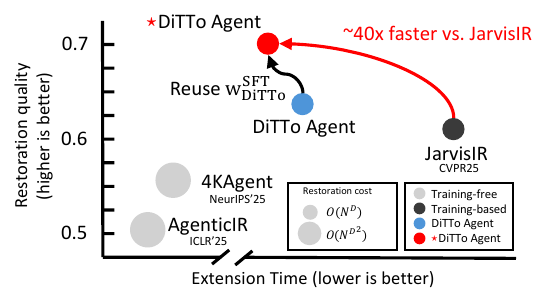}
        \subcaption{Quality vs.\ adaptation cost.}
        \label{fig:bubble_a}
    \end{minipage}
    \hfill
    \begin{minipage}{0.59\linewidth}
        \centering
        \includegraphics[width=\linewidth]{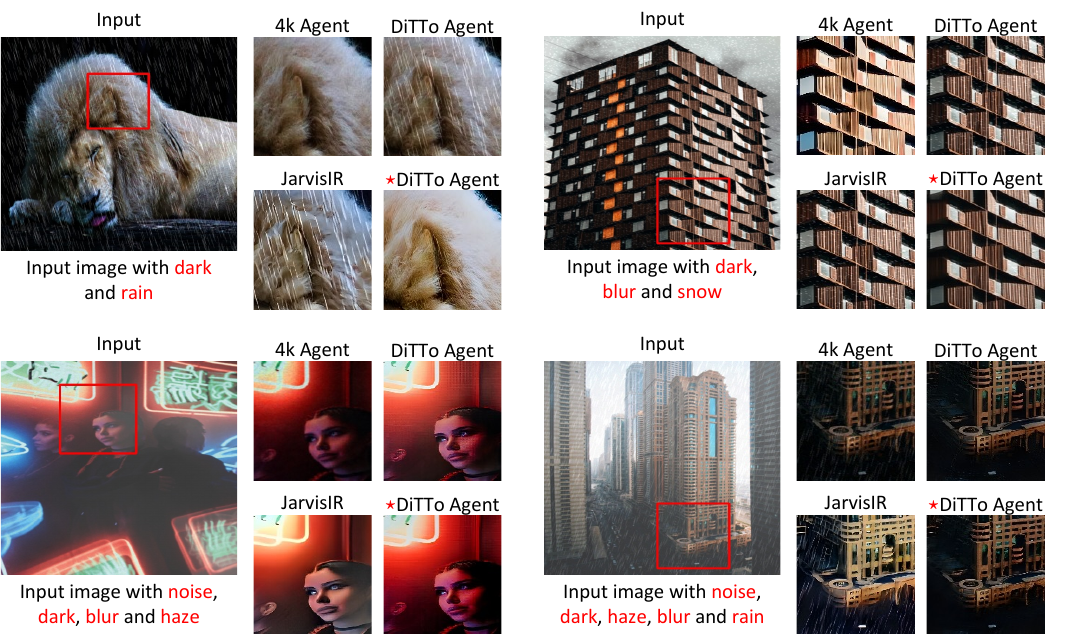}
        \subcaption{Qualitative comparison.}
        \label{fig:bubble_b}
    \end{minipage}
    \caption{
    \textbf{All-in-One (AiO) image restoration (IR) quality.}
    \textbf{(a)} Reusing the our pre-trained DiTTo SFT checkpoint enables $\sim$40$\times$ faster adaptation with higher IR quality than \textit{training-based} JarvisIR~\cite{lin2025jarvisir} when adding a new restoration-expert, showing stronger \textit{plug-and-play scalable extensibility}.
    \textbf{(b)} DiTTo Agent removes multi-degradations more thoroughly than prior agent-based AiO IR methods.}
    \label{fig:bubble}
\end{figure}

\section{Introduction}
\label{sec}
\vspace{-3mm}

Real-world images rarely suffer from a single degradation~\cite{11123156,10680296,10056934}.
Photographs taken outdoors are often affected simultaneously by fog, rain, and sensor noise~\cite{he2010single,11123156,10767188}, while indoor or handheld photography commonly combines motion blur and defocus blur with compression artifacts~\cite{nah2017deep,nah2021ntire}.
Despite this, most image restoration (IR) methods assume a single known degradation type~\cite{11123156,10680296,zamir2022restormer}, and all-in-one approaches~\cite{conde2024instructir,kong2024mioir,li2022airnet,potlapalli2023promptir}, although more general, often sacrifice peak performance for broader coverage.
As shown in Fig.~\ref{fig:bubble_b}, our DiTTo Agent removes multi-degradations more thoroughly than prior agent-based all-in-one IR methods.

More fundamentally, these methods overlook a critical insight: \textit{the order in which degradations are removed matters}~\cite{lin2024diffbir,10.1145/3532625}.
For example, applying low-light enhancement before de-noising can amplify noise, while performing de-fogging before de-raining can alter the apparent distribution of rain~\cite{kong2024mioir,10680296}.
This motivates an order-aware restoration paradigm that explicitly reasons about the restoration-action-ordering rather than treating degradations as simultaneously addressable.
This combinatorial nature naturally motivates \textit{agentic IR}, in which a vision-language model (VLM) leverages a pool of pre-built restoration-experts and \textit{sequentially} selects which restoration-expert to invoke at each step via structured tool calls, rather than learning a single monolithic restoration network itself.

\begin{figure*}[h]
\centering
\includegraphics[width=\linewidth]{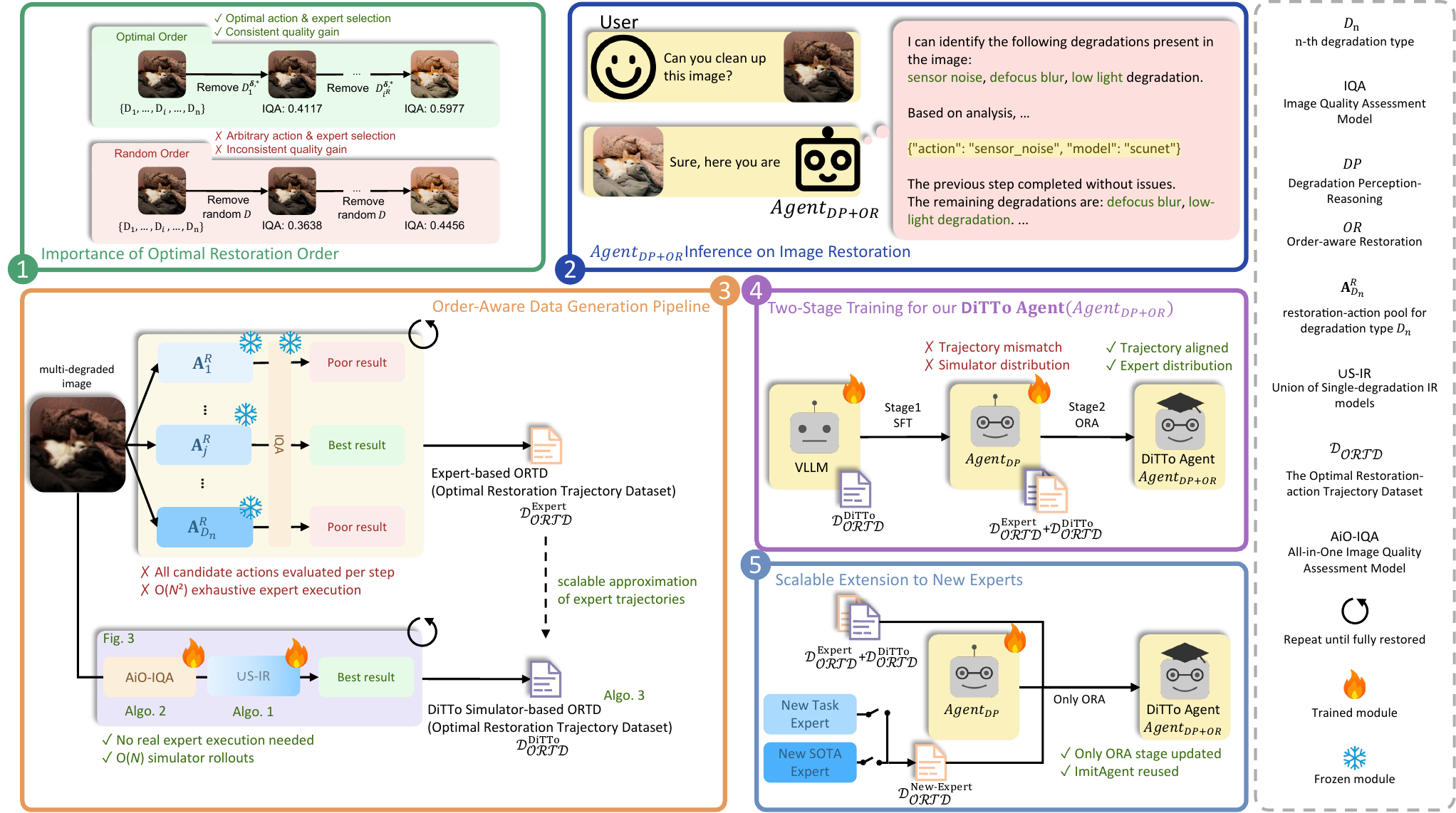}
\caption{
Overview of \textbf{DiTTo}, an order-aware image restoration agent framework.
\textbf{(1)} The same involved degradation type set can yield substantially different IQA scores depending on the restoration-action-ordering, motivating order-aware restoration.
\textbf{(2)} At inference time, $\mathit{Agent}_{DP+OR}$ identifies the involved degradation types and sequentially invokes restoration-experts via JSON-based tool calls.
\textbf{(3)} Prior training-based agents construct $\mathcal{D}_{\text{ORTD}}^{\text{Expert}}$ by evaluating all candidate restoration-actions with real restoration-experts at each restoration index, requiring $\mathcal{O}((N^{\mathbf{D}})^2)$ real restoration-expert calls per image.
The \textbf{DiTTo Simulator} instead constructs $\mathcal{D}_{\text{ORTD}}^{\text{DiTTo}}$ with $\mathcal{O}(N^{\mathbf{D}})$ simulator steps by using AiO-IQA to select the highest-scoring restoration-action identifier and $\cup$S-IR to produce the next restored image-state.
\textbf{(4)} The \textbf{DiTTo Agent} is trained in two stages: SFT on $\mathcal{D}_{\text{ORTD}}^{\text{DiTTo}}$ yields checkpoint $W_{\text{DiTTo}}^{\text{SFT}}$, followed by DPO-based \textbf{Order-aware Restoration Alignment (ORA)} on a small subset of $\mathcal{D}_{\text{ORTD}}^{\text{Expert}}$ to align simulator-generated restoration-action-trajectories with expert-executed restoration-action-trajectories, producing final checkpoint $W_{\text{DiTTo}}^{\text{ORA}}$.
\textbf{(5)} When a new restoration-expert is added, DiTTo reuses $\cup$S-IR, AiO-IQA, and $W_{\text{DiTTo}}^{\text{SFT}}$, and updates only the efficient ORA stage.
}
\label{fig:teaser}
\end{figure*}

We formalize this setting using the notation in \textit{Appendix}.~\ref{sec:notation}.
Let $\mathbf{D}=\{D_1,\ldots,D_{N^{\mathbf{D}}}\}$ denote a predefined universe of $N^{\mathbf{D}}$ degradation types.
Given a clean image-state $I_{\text{clean}}$, a degradation-ordering $\boldsymbol{\delta}$ induces a degradation-action-trajectory $\mathcal{T}^{\boldsymbol{\delta}}$ and produces the observed multi-degraded input $I^{\boldsymbol{\delta}}_j$.
While $\mathcal{T}^{\boldsymbol{\delta}}$ is known by construction during synthetic data generation, the inverse optimal restoration-action-trajectory $\mathcal{T}^{\boldsymbol{\delta},*}$ must be discovered.
At restoration index $i^R$, the optimal step is denoted by $\rho^{\boldsymbol{\delta},*}_{i^R}$, and the Optimal Restoration-action Trajectory Dataset (ORTD) consists of restored image-state ($\widetilde{I}^{\boldsymbol{\delta},*}_{i^R}$) and optimal restoration-action ($A^R_{\rho^{\boldsymbol{\delta},*}_{i^R}}$) pairs $(\widetilde{I}^{\boldsymbol{\delta},*}_{i^R}, A^R_{\rho^{\boldsymbol{\delta},*}_{i^R}})$ along $\mathcal{T}^{\boldsymbol{\delta},*}$.
As illustrated in Fig.~\ref{fig:teaser}.\textcircled{1}, applying restoration-actions in a suboptimal restoration-action-ordering leads to measurably lower IQA scores at intermediate restored image-states, causing early errors to propagate into the final output.
This challenge has motivated agentic IR~\cite{chen2024restoreagent,jiang2025multi,zhu2025agenticir}, where an agent sequentially selects restoration-actions realized by restoration-experts to maximize final restoration quality.

A fully capable restoration agent must address two complementary challenges: first, identifying which degradations are present in the input, a capability we term \textit{Degradation Perception-Reasoning} ($\mathit{Agent}_{DP}$); and second, determining their optimal restoration-action-ordering $\boldsymbol{\rho}^{\boldsymbol{\delta},*}$, which we term \textit{Order-aware Restoration} ($\mathit{Agent}_{OR}$).
As shown in Fig.~\ref{fig:teaser}.\textcircled{2}, the target system is the composition $\mathit{Agent}_{DP+OR}$, which jointly performs both functions: it reasons over the detected involved type set $\mathbf{D}^{\boldsymbol{\delta}}\subseteq\mathbf{D}$ and sequentially invokes restoration-actions $A^R_{\rho^{\boldsymbol{\delta},*}_{i^R}}$ at each restoration index $i^R$ via structured tool calls to progressively restore the image.
Recent agentic IR systems~\cite{chen2024restoreagent,zhu2025agenticir,jiang2025multi} similarly aim toward this combined capability.

Although $\mathcal{T}^{\boldsymbol{\delta}}$ is known when synthesizing training data, its inverse optimal restoration-action-trajectory $\mathcal{T}^{\boldsymbol{\delta},*}$ is not directly given: at each $\widetilde{I}^{\boldsymbol{\delta}}_{i^R}$, identifying $\rho^{\boldsymbol{\delta},*}_{i^R}$ requires evaluating every candidate identifier $\rho\in\mathbf{A}^{\boldsymbol{\delta}}_{i^R}$, i.e., applying the corresponding restoration-action $A^R_{\rho}(\cdot)$ to the current restored image-state and IQA-scoring the resulting next restored image-state.
Training-free agents perform this search at inference time, incurring $\mathcal{O}((N^{\mathbf{D}})^{2})$ real restoration-expert calls per image and often producing suboptimal restoration-action-trajectories when the search fails to faithfully track real restoration-expert behavior~\cite{zhu2025agenticir,jiang2025multi}.
Training-based agents avoid this inference-time search by transferring it to offline ORTD construction, training a policy on ground-truth image-state and optimal restoration-action pairs and reducing inference to $\mathcal{O}(N^{\mathbf{D}})$ real restoration-expert calls.
However, existing training-based approaches still require expensive ORTD construction.
Each ground-truth pair $(\widetilde{I}^{\boldsymbol{\delta},*}_{i^R}, A^R_{\rho^{\boldsymbol{\delta},*}_{i^R}})$ requires evaluating all candidates in $\mathbf{A}^{\boldsymbol{\delta}}_{i^R}$ with real restoration-experts~\cite{chen2024restoreagent,lin2025jarvisir,zhou2025qagent} and IQA-scoring their resulting next restored image-states, yielding $\mathcal{O}((N^{\mathbf{D}})^{2})$ real restoration-expert calls per image for offline ORTD pair generation.
Moreover, because ORTD pair generation is coupled with real restoration-expert calls, \textit{adding a new restoration-expert requires regenerating expert-based ORTD pairs and retraining the agent}, making these approaches inherently non-modular, which limits their scalability as the restoration-expert pool grows (overview of the full ORTD pair generation is provided in \textit{Appendix} Fig.~\ref{fig:degradation_process}).

To address the two bottlenecks of costly offline ORTD pair generation and non-modular restoration-expert extension, we propose \textbf{DiTTo}, an order-aware image restoration agent framework that leverages a learned simulator and a pool of pre-built restoration-experts to construct supervision and efficiently select restoration-actions without coupling agent training to real restoration-expert calls.
This design targets two practical requirements for deployable restoration agents: scalable supervision construction and efficient adaptation to an expanding restoration-expert pool.
DiTTo consists of the \textbf{DiTTo Simulator}, which constructs $\mathcal{D}_{\text{ORTD}}^{\text{DiTTo}}$ using $\cup$S-IR and AiO-IQA, and the \textbf{DiTTo Agent}, which uses these simulator-generated ORTD pairs to learn $\mathit{Agent}_{DP+OR}$ through two-stage training.

\textbf{DiTTo Simulator.}
As illustrated in Fig.~\ref{fig:teaser}.\textcircled{3}, the DiTTo Simulator replaces $\mathcal{O}((N^{\mathbf{D}})^{2})$ real restoration-expert calls with $\mathcal{O}(N^{\mathbf{D}})$ single-step simulator steps.
We denote ORTDs constructed by real restoration-experts as $\mathcal{D}_{\text{ORTD}}^{\text{Expert}}$ and those approximated by our proposed simulator as $\mathcal{D}_{\text{ORTD}}^{\text{DiTTo}}$.
The simulator combines $\cup$\textbf{S-IR}, which approximates the single-restoration effect of a restoration-action, and \textbf{AiO-IQA}, which predicts the quality of the next restored image-state induced by each candidate restoration-action identifier.
At each restoration index $i^R$, AiO-IQA selects the highest-scoring candidate as the simulator-approximated optimal step $\rho^{\boldsymbol{\delta},*}_{i^R}$, and $\cup$S-IR applies the corresponding restoration-action to produce the next restored image-state.
The $\cup$S-IR is built on a novel \textit{adaptive frequency-band mixing} mechanism that performs action-conditioned mixing between clean-conditioned and degraded-conditioned features across separate frequency bands.

\textbf{DiTTo Agent.}
Building on $\mathcal{D}_{\text{ORTD}}^{\text{DiTTo}}$, the DiTTo Agent realizes $\mathit{Agent}_{DP+OR}$ through two-stage training.
Stage~1 trains a vision-language model (VLM) via supervised fine-tuning (SFT) on $\mathcal{D}_{\text{ORTD}}^{\text{DiTTo}}$, yielding the DiTTo-SFT checkpoint weight $W_{\text{DiTTo}}^{\text{SFT}}$.
Since simulator-generated restoration-action-trajectories may diverge from expert-executed restoration-action-trajectories, Stage~2 applies efficient \textbf{Order-aware Restoration Alignment (ORA)} using Direct Preference Optimization (DPO)~\cite{rafailov2023direct} with a small subset of $\mathcal{D}_{\text{ORTD}}^{\text{Expert}}$.
This produces the final $\mathit{Agent}_{DP+OR}$, our \textbf{DiTTo Agent}, with final checkpoint weight $W_{\text{DiTTo}}^{\text{ORA}}$.
When a new restoration-expert is added, DiTTo can reuse $\cup$S-IR, AiO-IQA, and $W_{\text{DiTTo}}^{\text{SFT}}$, and quickly updates $W_{\text{DiTTo}}^{\text{ORA}}$ only with efficient ORA stage.
As shown in Fig.~\ref{fig:bubble_a}, this enables $\sim$40$\times$ faster adaptation with higher IR quality than training-based JarvisIR~\cite{lin2025jarvisir} when a new restoration-expert is added.





\textbf{Contributions.}
\begin{itemize}
\item We propose \textbf{DiTTo}, a novel \textit{order-aware image restoration agent framework} that decouples agent training from real restoration-expert calls through a learned simulator, enabling scalable ORTD construction and plug-and-play restoration-expert extensibility.

\item We introduce \textbf{DiTTo Simulator}, which combines $\cup$\textbf{S-IR} for single-degradation restoration-action simulation and \textbf{AiO-IQA} for IQA-based next-state scoring, reducing ORTD pair generation from $\mathcal{O}((N^{\mathbf{D}})^{2})$ real restoration-expert calls to $\mathcal{O}(N^{\mathbf{D}})$ simulator steps per image.

\item We propose \textbf{DiTTo Agent}, trained by large-scale SFT on our efficiently constructed $\mathcal{D}_{\text{ORTD}}^{\text{DiTTo}}$ followed by efficient DPO-based \textit{Order-aware Restoration Alignment (ORA)} on a small subset of $\mathcal{D}_{\text{ORTD}}^{\text{Expert}}$, enabling fast plug-and-play adaptation to new restoration-experts.

\item Extensive experiments show that our DiTTo Agent achieves state-of-the-art multi-degradation restoration quality, and is the only agentic IR framework that simultaneously supports $\mathcal{O}(N^{\mathbf{D}})$ ORTD pair generation and efficient plug-and-play restoration-expert extensibility.
\end{itemize}

\newcommand{\yes}{\checkmark}
\newcommand{\no}{\times}
\newcommand{\na}{\text{N/A}}
\vspace{-5mm}
\begin{table}[h]
\centering
\caption{Comparison of image restoration approaches. Here, $N^{\mathbf{D}}=|\mathbf{D}|$ denotes the number of degradation types in the predefined degradation universe, and ORTD denotes the Optimal Restoration-action Trajectory Dataset used to supervise order-aware restoration agents.}
\label{tab:comparison}
\renewcommand{\arraystretch}{1.3}

\resizebox{0.6\linewidth}{!}{
\begin{tabular}{lcccc}
\toprule
& & \multicolumn{3}{c}{\textbf{Agent}} \\
\cmidrule(lr){3-5}
& \textbf{All-in-one} & \textbf{Training-free} & \textbf{Training-based} & \textbf{DiTTo (ours)} \\
\midrule
Order-aware restoration        & $\no$        & $\triangle$ & $\yes$        & $\yes$ \\
Restoration inference cost                & $\mathcal{O}(1)$ & $\mathcal{O}((N^{\mathbf{D}})^2)$ & $\mathcal{O}(N^{\mathbf{D}})$ & $\mathcal{O}(N^{\mathbf{D}})$ \\
ORTD pair generation cost & --      & --         & $\mathcal{O}((N^{\mathbf{D}})^2)$ & $\mathcal{O}(N^{\mathbf{D}})$ \\
Plug-and-play extensibility   & $\no$        & $\triangle$ & $\no$         & $\yes$ \\
\bottomrule
\end{tabular}
}

\vspace{4pt}
\footnotesize
$\yes$: Supported \quad
$\triangle$: Partial (fixed heuristics) \quad
$\no$: Not supported \quad
--: Not applicable
\end{table}

\paragraph{Related Works.}Tab.~\ref{tab:comparison} summarizes how DiTTo differs from prior multi-degradation IR paradigms in terms of order-aware restoration, inference cost, ORTD pair generation cost, and plug-and-play extensibility; a detailed discussion of Related Works is deferred to \textit{Appendix}.~\ref{sec:rw}.
\vspace{-3mm}
\section{Method}
\label{sec:method}

\subsection{Overview}
\label{sec:method_overview}

As introduced in Sec.~\ref{sec}, the overall \textbf{DiTTo} framework (Fig.~\ref{fig:teaser}) consists of two components: the \textbf{DiTTo Simulator} and the \textbf{DiTTo Agent}.
The DiTTo Simulator scalably constructs $\mathcal{D}_{\text{ORTD}}^{\text{DiTTo}}$ via two modules:
$\cup$\textbf{S-IR}, which instantiates the single-step simulator $\mathcal{S}_\theta$, and \textbf{AiO-IQA}, which instantiates the IQA-based scoring model $f_{\psi}$ over candidate identifiers in $\mathbf{A}^{\boldsymbol{\delta}}_{i^R}$.

The DiTTo Agent is then trained in two stages:
Stage~1 performs SFT on $\mathcal{D}_{\text{ORTD}}^{\text{DiTTo}}$ to obtain the DiTTo-SFT checkpoint $W_{\text{DiTTo}}^{\text{SFT}}$, and Stage~2 applies \textbf{Order-aware Restoration Alignment (ORA)} via DPO~\cite{rafailov2023direct} on a small subset of $\mathcal{D}_{\text{ORTD}}^{\text{Expert}}$ to obtain the final $\mathit{Agent}_{DP+OR}$ checkpoint $W_{\text{DiTTo}}^{\text{ORA}}$, while $\cup$S-IR, AiO-IQA, and $W_{\text{DiTTo}}^{\text{SFT}}$ remain reusable for plug-and-play restoration-expert adaptation.

\vspace{-3mm}
\paragraph{Problem Formulation}

Following the notation in Sec.~\ref{sec:notation} and Sec.~\ref{sec}, we formulate multi-degradation IR as a sequential decision-making problem for order-aware restoration.
Given an observed multi-degraded image-state $I^{\boldsymbol{\delta}}_j$ produced by a hidden degradation-ordering $\boldsymbol{\delta}$ over $\mathbf{D}^{\boldsymbol{\delta}}\subseteq\mathbf{D}$, the restoration-action-trajectory $\mathcal{T}^{\boldsymbol{\delta},\boldsymbol{\rho}^{\boldsymbol{\delta}}}$ removes the $j=|\mathbf{D}^{\boldsymbol{\delta}}|$ degradations step by step ($i^R=j,j{-}1,\ldots,1$).

At restoration index $i^R$, a step is specified by an identifier
$\rho^{\boldsymbol{\delta}}_{i^R}=(D^{\boldsymbol{\delta}}_{i^R}, i^E_{D^{\boldsymbol{\delta}}_{i^R}})\in\mathbf{A}^{\boldsymbol{\delta}}_{i^R}$, and the corresponding restoration-action $A^R_{\rho^{\boldsymbol{\delta}}_{i^R}}(\cdot)$ produces
$\widetilde{I}^{\boldsymbol{\delta}}_{i^R-1}=A^R_{\rho^{\boldsymbol{\delta}}_{i^R}}(\widetilde{I}^{\boldsymbol{\delta}}_{i^R})$, removing $D^{\boldsymbol{\delta}}_{i^R}$ from the remaining-type set and determining the next candidate set $\mathbf{A}^{\boldsymbol{\delta}}_{i^R-1}$.
Given an image-quality score $Q$ (higher is better) and the set $\mathcal{P}^{\boldsymbol{\delta}}$ of all valid restoration-action-orderings, the optimal restoration-action-ordering
$\boldsymbol{\rho}^{\boldsymbol{\delta},*}=(\rho^{\boldsymbol{\delta},*}_{i^R})_{i^R=j}^{1}$
is the element of $\mathcal{P}^{\boldsymbol{\delta}}$ whose induced $\mathcal{T}^{\boldsymbol{\delta},*}$ attains the highest $Q$ on $\widetilde{I}^{\boldsymbol{\delta},*}_0$, and $\mathcal{T}^{\boldsymbol{\delta},*}$ induces $\mathcal{D}_{\text{ORTD}}=\bigcup_{\boldsymbol{\delta}}\{(\widetilde{I}^{\boldsymbol{\delta},*}_{i^R}, A^R_{\rho^{\boldsymbol{\delta},*}_{i^R}}(\cdot))\}_{i^R=j}^{1}$.

\vspace{-3mm}

As discussed in Sec.~\ref{sec}, constructing $\mathcal{D}_{\text{ORTD}}^{\text{Expert}}$ directly costs $\mathcal{O}((N^{\mathbf{D}})^2)$ real restoration-expert calls per image and couples ORTD pair generation to a fixed restoration-expert set.
DiTTo addresses these two bottlenecks by constructing a scalable $\mathcal{D}_{\text{ORTD}}^{\text{DiTTo}}$ with the DiTTo Simulator and using a small subset of $\mathcal{D}_{\text{ORTD}}^{\text{Expert}}$ only for our proposed ORA.

\subsection{DiTTo Simulator}
\label{sec:method_sim}

\begin{figure}[h]
    \centering
    \includegraphics[width=1\linewidth]{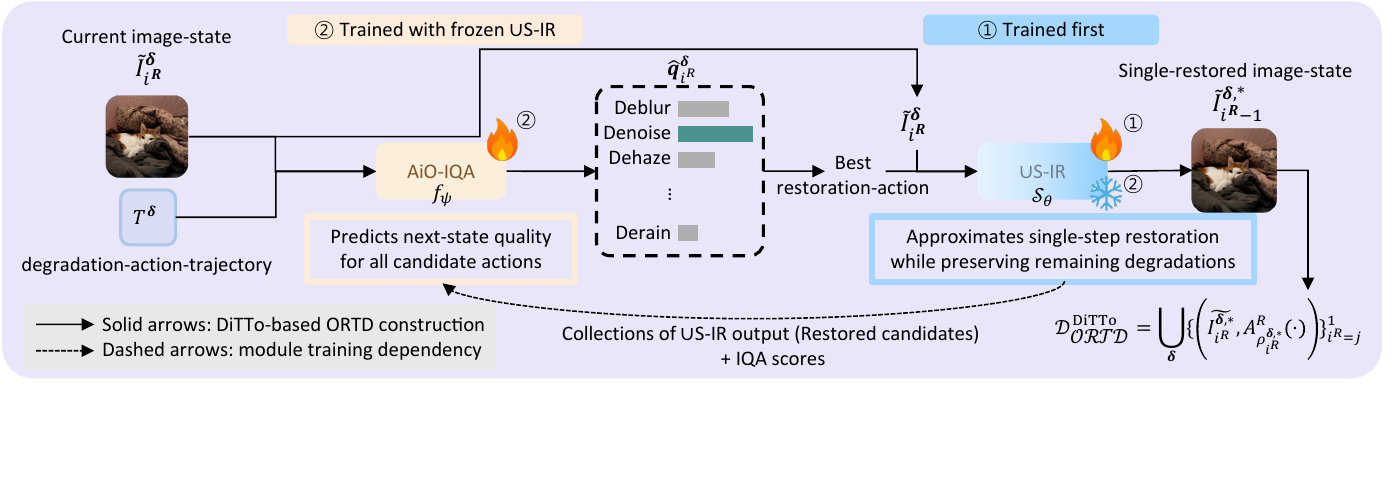}
    \vspace{-15mm}
\caption{
Overview of the \textbf{DiTTo Simulator}, which constructs $\mathcal{D}_{\text{ORTD}}^{\text{DiTTo}}$ without exhaustive real restoration-expert calls.
The simulator consists of two modules: $\cup$\textbf{S-IR}, instantiated as the single-degradation restoration simulator $\mathcal{S}_{\theta}$, and \textbf{AiO-IQA}, instantiated as the IQA-based scoring model $f_{\psi}$.
$\cup$S-IR is first trained to approximate the next restored image-state induced by a candidate restoration-action identifier $\rho\in\mathbf{A}^{\boldsymbol{\delta}}_{i^R}$, and is then frozen when training AiO-IQA.
Given the current restored image-state $\widetilde{I}^{\boldsymbol{\delta}}_{i^R}$ and the degradation-action-trajectory $\mathcal{T}^{\boldsymbol{\delta}}$, AiO-IQA predicts per-action quality scores $\hat{\mathbf{q}}^{\boldsymbol{\delta}}_{i^R}$ over $\mathbf{A}^{\boldsymbol{\delta}}_{i^R}$.
The highest-scoring identifier $\rho^{\boldsymbol{\delta}}_{i^R}$ is selected, and $\mathcal{S}_{\theta}$ applies the corresponding restoration-action to produce $\widetilde{I}^{\boldsymbol{\delta}}_{i^R-1}$.
Repeating this procedure from $i^R=j$ to $1$ constructs $\mathcal{D}_{\text{ORTD}}^{\text{DiTTo}}$ with only $\mathcal{O}(N^{\mathbf{D}})$ simulator steps per image.
The training of $\cup$S-IR, training of AiO-IQA with frozen $\cup$S-IR, and construction of $\mathcal{D}_{\text{ORTD}}^{\text{DiTTo}}$ are detailed in Algorithms~\ref{alg:train_usir}, \ref{alg:train_aio_iqa}, and~\ref{alg:construct_ditto_ortd}, respectively.
}
    \label{fig:ditto_simulator}
\end{figure}

As shown in Fig.~\ref{fig:ditto_simulator}, the \textbf{DiTTo Simulator} consists of two complementary modules, $\cup$\textbf{S-IR} and \textbf{AiO-IQA}.
Together, these modules construct $\mathcal{D}_{\text{ORTD}}^{\text{DiTTo}}$ with only $\mathcal{O}(N^{\mathbf{D}})$ simulator steps per image, avoiding exhaustive real restoration-expert calls.

\subsubsection{\texorpdfstring{$\cup$S-IR: Single-degradation Restoration Simulator}{US-IR: Single-degradation Restoration Simulator}}
\label{sec:method_usir}

The purpose of $\cup$S-IR is not to serve as the final restoration model, but to provide a cheap single-restoration approximation of heterogeneous restoration-experts.
Given a current restored image-state $\widetilde{I}^{\boldsymbol{\delta}}_{i^R}$ and a candidate restoration-action identifier $\rho\in\mathbf{A}^{\boldsymbol{\delta}}_{i^R}$, $\cup$S-IR predicts
$
\widehat{I}^{\boldsymbol{\delta}}_{i^R-1,\rho}
=
\mathcal{S}_{\theta}
\!\left(
\widetilde{I}^{\boldsymbol{\delta}}_{i^R},
\rho
\right),
$
where the degradation type specified by $\rho$ is removed while the other remaining degradations are preserved.
This single-restoration is crucial for training AiO-IQA: obtaining candidate scores with real restoration-experts would require applying all candidates at each restoration index, reintroducing the $\mathcal{O}((N^{\mathbf{D}})^2)$ real expert cost.
By replacing these expert calls with inexpensive simulator steps, $\cup$S-IR makes large-scale candidate supervision feasible.

We implement $\cup$S-IR with action-conditioned clean/degraded feature mixing, where clean-conditioned features provide restoration cues for the target degradation and degraded-conditioned features help preserve non-target degradations.
The architecture, conditioning scheme, frequency-band selective gating, and training objective of $\cup$S-IR are provided in \textit{Appendix}.~\ref{sec:usir_details}.

\subsubsection{AiO-IQA: All-in-One Restoration-Action Scoring}
\label{sec:method_iqa}

Although $\cup$S-IR makes single-restoration simulation cheap, using it to evaluate all candidate restoration-actions at every restoration index would still be inefficient for large-scale ORTD construction.
AiO-IQA removes this remaining bottleneck by directly predicting per-action quality scores from the current restored image-state and the degradation-action-trajectory of the instance.
Given $\widetilde{I}^{\boldsymbol{\delta}}_{i^R}$, $\mathcal{T}^{\boldsymbol{\delta}}$, and $\mathbf{A}^{\boldsymbol{\delta}}_{i^R}$, AiO-IQA outputs
$
\hat{\mathbf{q}}^{\boldsymbol{\delta}}_{i^R}
=
f_{\psi}
\!\left(
\widetilde{I}^{\boldsymbol{\delta}}_{i^R},
\mathcal{T}^{\boldsymbol{\delta}}
\right)
\in {[0,1]}^{|\mathbf{A}^{\boldsymbol{\delta}}_{i^R}|}
,
$
where each entry is aligned with one candidate restoration-action in
$\mathbf{A}^{\boldsymbol{\delta}}_{i^R}$ under a fixed enumeration and estimates the quality of the next restored image-state induced by the corresponding restoration-action $A^R_{\rho}(\cdot)$.
The simulator-approximated optimal step is selected as
$
\rho^{\boldsymbol{\delta},*}_{i^R}
=
\operatorname*{argmax}_{\rho\in\mathbf{A}^{\boldsymbol{\delta}}_{i^R}}
\hat{\mathbf{q}}^{\boldsymbol{\delta}}_{i^R}[\rho].
$
Only the corresponding restoration-action $A^R_{\rho^{\boldsymbol{\delta},*}_{i^R}}$ is applied through $\cup$S-IR to produce the next restored image-state.
Repeating this procedure from $i^R=j$ to $1$ constructs one DiTTo-based restoration-action-trajectory using $\mathcal{O}(N^{\mathbf{D}})$ simulator steps per image.

AiO-IQA is trained to approximate the IQA-based ranking of candidate restoration-actions.
For supervision, $\cup$S-IR first generates one-step candidate next states, which are scored by a fixed ensemble of full-reference and no-reference IQA metrics.
The model then learns to predict the relative quality of candidate restoration-actions directly from the current restored image-state and the degradation-action-trajectory $\mathcal{T}^{\boldsymbol{\delta}}$.
The metric ensemble, score normalization, ranking objective, and architecture details of $f_{\psi}$ are provided in \textit{Appendix}.~\ref{sec:aioiqa_details}.

\subsection{Two-Stage DiTTo Agent Training}
\label{sec:method_agent}

The \textbf{DiTTo Agent} is a VLM trained to perform Degradation Perception-Reasoning, Order-aware Restoration, and structured tool use, leveraging a pool of restoration-experts rather than learning a restoration network itself.
It receives the observed multi-degraded image-state and produces a sequence of restoration-action tool calls, each specifying which degradation type to remove and which restoration-expert to invoke.
We train the agent with Stage~1 SFT and Stage~2 ORA.
\vspace{-1.5mm}
\paragraph{Stage~1: SFT on DiTTo-based ORTD.}
Using the trained DiTTo Simulator, we construct $\mathcal{D}_{\text{ORTD}}^{\text{DiTTo}}$ by the procedure described in Sec.~\ref{sec:method_sim}.
The resulting image-state and optimal restoration-action pairs are converted into multi-turn tool-use conversations, in which the agent reasons about the current degradation state and produces a structured restoration-action call as a JSON-based tool call (\textit{Appendix}.~\ref{sec:agent_details}).
SFT on these conversations yields the DiTTo-SFT checkpoint $W_{\text{DiTTo}}^{\text{SFT}}$, which acquires Degradation Perception-Reasoning, Order-aware Restoration, and structured tool use.
\vspace{-1.5mm}
\paragraph{Stage~2: Order-aware Restoration Alignment (ORA).}
Stage~1 is supervised by simulator-generated restoration-action-trajectories from $\mathcal{D}_{\text{ORTD}}^{\text{DiTTo}}$, whereas deployment relies on real restoration-experts whose single-restoration may differ from the simulator $\mathcal{S}_{\theta}$.
To reduce this simulator-to-expert gap, Stage~2 refines the agent's \emph{planning} ability for order-aware restoration, including degradation perception, restoration-action-ordering, restoration-expert selection, and JSON-based tool-call validity.

A standard DPO objective treats each restoration-action-trajectory response as a single \emph{chosen}-versus-\emph{rejected} sequence and optimizes one response-level preference margin.
However, in our setting, chosen and rejected responses often share most reasoning and JSON-template tokens, while differing only in trajectory-relevant parts, such as $\mathbf{D}^{\boldsymbol{\delta}}$, $\boldsymbol{\pi}^{\boldsymbol{\delta}}$, $\pi^{\boldsymbol{\delta}}_{i^R}$, and JSON-based tool-call validity.
This can dilute the preference signal needed for order-aware restoration.

To overcome this, ORA computes preference margins over decomposed planning axes rather than over the entire response at once.
Specifically, we decompose each response into DP, OR, and tool-call axes, corresponding to Degradation Perception-Reasoning, Order-aware Restoration, and JSON-based tool use, respectively, and align chosen and rejected responses axis-wise.
We construct a small expert-executed ORTD subset $\mathcal{D}_{\text{ORTD}}^{\text{Expert}}$ as chosen trajectories and pair them with simulator-generated or perturbed rejected trajectories for DPO-based ORA.
The detailed DPO formulations, decomposed axes construction, weighting scheme for decomposed axes, and rejected-response variants are provided in \textit{Appendix}.~\ref{sec:agent_details}.
\vspace{-1.5mm}
\paragraph{Plug-and-Play Restoration-Expert Extensibility.}
The two-stage design decouples large-scale agent training from the concrete real restoration-expert set.
When a new restoration-expert is added, DiTTo can reuse $\cup$S-IR, AiO-IQA, and $W_{\text{DiTTo}}^{\text{SFT}}$, and updates only the efficient ORA stage with a small Expert-based ORTD subset involving the new restoration-expert.
Thus, DiTTo avoids rerunning the full ORTD pair generation pipeline, unlike prior training-based agents whose supervision is tied to a fixed restoration-expert set.
\section{Experiments}
\label{sec:experiments}

The restoration-expert pool, training data construction, and implementation details (VLM backbone, LoRA configuration, optimizer, and training schedule for $\cup$S-IR, AiO-IQA, SFT, and ORA) are provided in \textit{Appendix}.

\begin{table*}[h]
\centering
\setlength{\belowcaptionskip}{-0.2cm}
\caption{
Quantitative comparison on the MiO-100 evaluation set with $j\in\{2,3,4,5\}$ concurrent degradations.
We report no-reference image-quality metrics on the final restored image-state $\widetilde{I}^{\boldsymbol{\delta},*}_0$.
\textbf{DiTTo Agent} uses the same restoration-expert pool as JarvisIR, while \scalebox{1.2}{\color{sh_red}{$\star$}}\textbf{DiTTo Agent} uses the extended restoration-expert pool.
We highlight the \colorbox{lightpeach!90}{best}, \colorbox{customblue}{second-best}, and \colorbox{custompink}{third-best} results.
}
\label{perception_table}
\resizebox{\textwidth}{!}{%
\setlength\tabcolsep{8pt}
\renewcommand\arraystretch{1}
\begin{tabular}{lcccccccc}
\toprule
 & \multicolumn{4}{c}{2 Degradations} & \multicolumn{4}{c}{3 Degradations} \\
\cmidrule(lr){2-5} \cmidrule(lr){6-9}
\multirow{-2}{*}{Method}
 & MUSIQ $\uparrow$ & MANIQA $\uparrow$ & CLIP-IQA+ $\uparrow$ & NIQE $\downarrow$
 & MUSIQ $\uparrow$ & MANIQA $\uparrow$ & CLIP-IQA+ $\uparrow$ & NIQE $\downarrow$ \\
\midrule
\multicolumn{9}{l}{\textit{All-in-One Methods}} \\
\midrule
AirNet~\cite{li2022airnet}
 & 59.85 & 0.3980 & 0.5410 & 8.323
 & 41.60 & 0.2887 & 0.4221 & 9.699 \\
PromptIR~\cite{potlapalli2023promptir}
 & 62.95 & 0.3810 & 0.5441 & 5.823
 & 53.38 & 0.2808 & 0.4078 & 6.162 \\
MiOIR~\cite{kong2024mioir}
 & 62.89 & 0.3906 & 0.5481 & 5.682
 & 52.41 & 0.2815 & 0.4040 & 6.075 \\
DA-CLIP~\cite{luo2024daclip}
 & 57.12 & 0.3536 & 0.5470 & 7.737
 & 46.78 & 0.2983 & 0.4714 & 9.016 \\
InstructIR~\cite{conde2024instructir}
 & 64.08 & 0.4139 & 0.4827 & 7.912
 & 46.53 & 0.2697 & 0.3440 & 9.221 \\
AutoDIR~\cite{jiang2024autodir}
 & 64.04 & 0.3667 & 0.5113 & 7.326
 & 52.29 & 0.3184 & 0.3985 & 8.537 \\
\midrule
\multicolumn{9}{l}{\textit{Agent-based Methods}} \\
\midrule
AgenticIR~\cite{zhu2025agenticir}
 & 63.10 & 0.5170 & 0.6595 & 6.038
 & 61.20 & 0.4585 & 0.6010 & 6.587 \\
4KAgent~\cite{zuo2026kagent}
 & 67.20 & 0.5640 & 0.7165 & 5.612
 & 65.40 & 0.5025 & 0.6555 & 6.121 \\
JarvisIR~\cite{lin2025jarvisir}
 & \colorbox{custompink}{69.34} & \colorbox{custompink}{0.5973} & \colorbox{custompink}{0.7464} & \colorbox{customblue}{5.366}
 & \colorbox{customblue}{67.54} & \colorbox{custompink}{0.5331} & \colorbox{custompink}{0.6845} & \colorbox{customblue}{5.862} \\
DiTTo Agent
 & \colorbox{customblue}{69.40} & \colorbox{customblue}{0.6457} & \colorbox{customblue}{0.7762} & \colorbox{custompink}{5.475}
 & \colorbox{custompink}{67.09} & \colorbox{customblue}{0.5823} & \colorbox{customblue}{0.7126} & \colorbox{custompink}{5.962} \\
\scalebox{1.5}{\color{sh_red}{$\star$}}\textbf{DiTTo Agent}
 & \colorbox{lightpeach!80}{71.76} & \colorbox{lightpeach!80}{0.7127} & \colorbox{lightpeach!80}{0.8393} & \colorbox{lightpeach!80}{5.208}
 & \colorbox{lightpeach!80}{70.65} & \colorbox{lightpeach!80}{0.6855} & \colorbox{lightpeach!80}{0.8101} & \colorbox{lightpeach!80}{5.773} \\
\midrule
 & \multicolumn{4}{c}{4 Degradations} & \multicolumn{4}{c}{5 Degradations} \\
\cmidrule(lr){2-5} \cmidrule(lr){6-9}
\multirow{-2}{*}{Method}
 & MUSIQ $\uparrow$ & MANIQA $\uparrow$ & CLIP-IQA+ $\uparrow$ & NIQE $\downarrow$
 & MUSIQ $\uparrow$ & MANIQA $\uparrow$ & CLIP-IQA+ $\uparrow$ & NIQE $\downarrow$ \\
\midrule
\multicolumn{9}{l}{\textit{All-in-One Methods}} \\
\midrule
AirNet~\cite{li2022airnet}
 & 42.98 & 0.3018 & 0.4361 & 9.946
 & 46.81 & 0.3341 & 0.4922 & 11.346 \\
PromptIR~\cite{potlapalli2023promptir}
 & 53.32 & 0.2721 & 0.4744 & 7.081
 & 58.20 & 0.3028 & 0.5515 & 6.527 \\
MiOIR~\cite{kong2024mioir}
 & 54.16 & 0.2795 & 0.4850 & 6.688
 & 58.99 & 0.3093 & 0.5528 & 6.430 \\
DA-CLIP~\cite{luo2024daclip}
 & 48.33 & 0.3119 & 0.4870 & 9.245
 & 52.63 & 0.3453 & 0.5495 & 10.547 \\
InstructIR~\cite{conde2024instructir}
 & 48.07 & 0.2820 & 0.3554 & 9.455
 & 52.35 & 0.3121 & 0.4011 & 10.787 \\
AutoDIR~\cite{jiang2024autodir}
 & 54.03 & 0.3329 & 0.4118 & 8.755
 & 58.83 & 0.3685 & 0.4646 & 9.988 \\
\midrule
\multicolumn{9}{l}{\textit{Agent-based Methods}} \\
\midrule
AgenticIR~\cite{zhu2025agenticir}
 & 63.50 & 0.5145 & 0.6635 & 5.864
 & 64.55 & 0.5325 & 0.6755 & 5.989 \\
4KAgent~\cite{zuo2026kagent}
 & 67.95 & 0.5615 & 0.7225 & 5.428
 & 69.10 & 0.5818 & 0.7355 & 5.547 \\
JarvisIR~\cite{lin2025jarvisir}
 & \colorbox{custompink}{70.25} & \colorbox{custompink}{0.5971} & \colorbox{custompink}{0.7535} & \colorbox{customblue}{5.167}
 & \colorbox{customblue}{71.36} & \colorbox{custompink}{0.6191} & \colorbox{custompink}{0.7677} & \colorbox{custompink}{5.292} \\
DiTTo Agent
 & \colorbox{customblue}{70.31} & \colorbox{customblue}{0.6402} & \colorbox{customblue}{0.7751} & \colorbox{custompink}{5.292}
 & \colorbox{custompink}{69.34} & \colorbox{customblue}{0.6270} & \colorbox{customblue}{0.7690} & \colorbox{customblue}{5.224} \\
\scalebox{1.5}{\color{sh_red}{$\star$}}\textbf{DiTTo Agent}
 & \colorbox{lightpeach!80}{71.84} & \colorbox{lightpeach!80}{0.7163} & \colorbox{lightpeach!80}{0.8443} & \colorbox{lightpeach!80}{5.160}
 & \colorbox{lightpeach!80}{72.27} & \colorbox{lightpeach!80}{0.7241} & \colorbox{lightpeach!80}{0.8509} & \colorbox{lightpeach!80}{5.188} \\
\bottomrule
\end{tabular}%
}
\end{table*}
\vspace{-4mm}

\paragraph{Evaluation Dataset.}
We evaluate all methods on the MiO-100 evaluation set introduced by MiOIR~\cite{kong2024mioir}.
Following the standard multi-degradation IR protocol~\cite{kong2024mioir,zhu2025agenticir}, we synthesize multi-degraded inputs by injecting $j \in \{2, 3, 4, 5\}$ degradations from the involved type set $\mathbf{D}$ into each clean image, where $\mathbf{D}$ covers six degradation types: sensor noise, low-light degradation, fog, defocus blur, rain streaks, and snow.
For each $j$, we sample six representative degradation-type combinations from $\mathbf{D}$ and apply each combination to all 100 clean images, yielding 600 multi-degraded test images per $j$ and 2{,}400 in total.
\vspace{-1.5mm}
\paragraph{Baselines and Restoration-Expert Pool.}
We compare \textbf{DiTTo Agent} against two categories of multi-degradation IR methods.
The first is \emph{All-in-One IR}: AirNet~\cite{li2022airnet}, PromptIR~\cite{potlapalli2023promptir}, MiOIR~\cite{kong2024mioir}, DA-CLIP~\cite{luo2024daclip}, InstructIR~\cite{conde2024instructir}, and AutoDIR~\cite{jiang2024autodir}.
The second is \emph{agent-based IR}: AgenticIR~\cite{zhu2025agenticir}, 4KAgent~\cite{zuo2026kagent}, and JarvisIR~\cite{lin2025jarvisir}.
For a fair head-to-head comparison, \textbf{DiTTo Agent} uses the same restoration-expert pool as JarvisIR~\cite{lin2025jarvisir}, while \scalebox{1.2}{\color{sh_red}{$\star$}}\textbf{DiTTo Agent} extends this pool with recent state-of-the-art restoration-experts to demonstrate plug-and-play scalable extensibility.
\vspace{-1.5mm}
\paragraph{Evaluation Metrics.}
We report four no-reference image-quality metrics on the final restored image-state $\widetilde{I}^{\boldsymbol{\delta},*}_0$: MUSIQ~\cite{ke2021musiq}, MANIQA~\cite{yang2022maniqa}, CLIP-IQA+~\cite{wang2023exploring}, and NIQE~\cite{mittal2012making}.
For each $j$, all numbers are averaged over the six combinations and 100 clean images per combination.

\subsection{Quantitative and Qualitative Comparison}
\label{sec:exp_comparison}

\paragraph{Quantitative results.}
Tab.~\ref{perception_table} shows that agent-based IR methods consistently outperform All-in-One methods across different numbers of concurrent degradations, confirming the advantage of sequentially invoking restoration-experts for multi-degradation restoration.
Among methods using the same restoration-expert pool, \textbf{DiTTo Agent} consistently outperforms previous agent-based IR methods on the final restored image-state $\widetilde{I}^{\boldsymbol{\delta},*}_0$.
This demonstrates that learning $\mathit{Agent}_{DP+OR}$ from simulator-generated ORTD pairs and further aligning it through Order-aware Restoration Alignment (ORA) yields more effective Order-aware Restoration and restoration-expert selection than existing agent-based IR pipelines.
With the extended restoration-expert pool, \scalebox{1.2}{\color{sh_red}{$\star$}}\textbf{DiTTo Agent} further improves restoration quality, validating the plug-and-play scalable extensibility of the proposed framework.
\paragraph{Qualitative results.}
\begin{figure}[h]
    \centering
    \includegraphics[width=\linewidth]{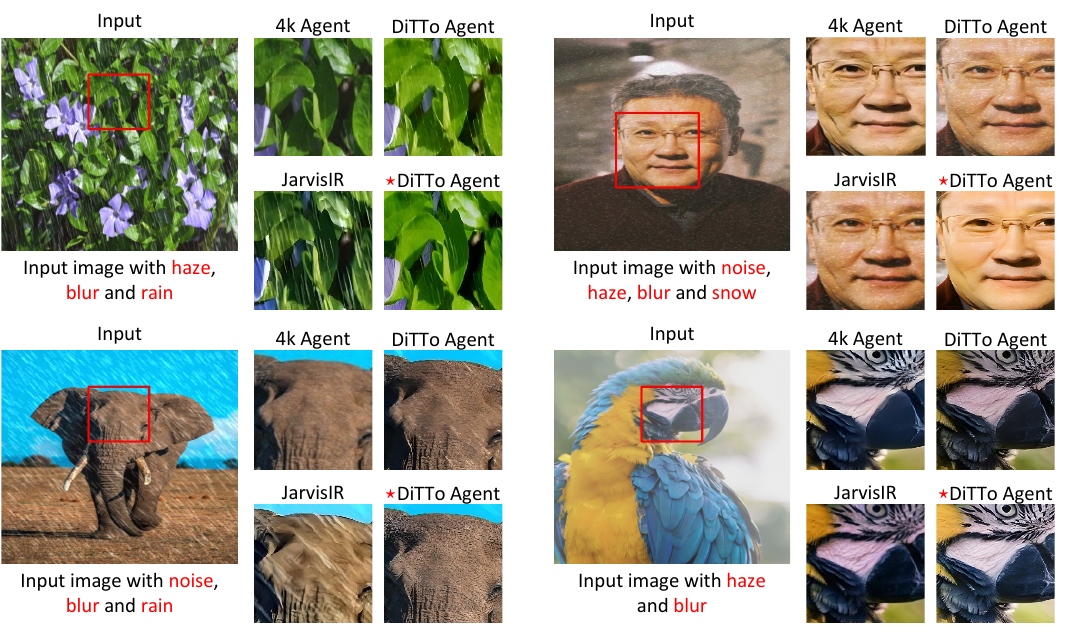}
    \caption{
Qualitative comparison on multi-degraded images shows that DiTTo Agent more effectively removes mixed degradations while preserving natural textures and semantic details.
}
    \label{fig:qualitative}
\end{figure}

Fig.~\ref{fig:qualitative} shows qualitative comparisons on challenging multi-degraded inputs containing haze, blur, rain, noise, and snow. Compared with 4K Agent and JarvisIR, DiTTo Agent more consistently removes mixed degradations while preserving semantic structures and fine textures, producing clearer details, more natural colors, and sharper object boundaries. These results suggest that the proposed order-aware restoration policy effectively improves perceptual restoration quality. Additional qualitative comparisons are provided in \textit{Appendix}.~\ref{sec:qualitative}.

\subsection{Ablation Studies}
\label{sec:exp_ablation}
\begin{table}[h]
\centering
\caption{
Ablation study on Order-aware Restoration Alignment (ORA).
\textbf{(a):} Effect of ORA compared with SFT-only and Generic DPO.
\textbf{(b):} Effect of each decomposed planning axis in ORA ((a).C).
DP, OR, and Tool denote Degradation Perception-Reasoning, Order-aware Restoration, and JSON-based tool-call format, respectively.
We report no-reference image-quality metrics on the final restored image-state $\widetilde{I}^{\boldsymbol{\delta},*}_0$.
We highlight the \colorbox{lightpeach!90}{best} and \colorbox{customblue}{second-best} results.
}
\label{tab:ablation}

\begin{minipage}[t]{0.48\linewidth}
\centering
\subcaption{Order-aware Restoration Alignment.}
\label{tab:ablation_ora}
\resizebox{\linewidth}{!}{%
\setlength\tabcolsep{4pt}
\renewcommand\arraystretch{1.1}
\begin{tabular}{clcccc}
\toprule
Variant & Method
 & MUSIQ $\uparrow$ & MANIQA $\uparrow$ & CLIP-IQA+ $\uparrow$ & NIQE $\downarrow$ \\
\midrule
A & $\mathit{Agent}_{DP}$ (SFT only)
 & 60.03 & 0.5083 & 0.5944 & 7.085 \\
B & $\mathit{Agent}_{DP}$ + Generic DPO~\cite{rafailov2023direct}
 & \colorbox{customblue}{67.21} & \colorbox{customblue}{0.5512} & \colorbox{customblue}{0.6984} & \colorbox{customblue}{5.157} \\
C & \cellcolor{lightpeach!30}$\mathit{Agent}_{DP+OR}$ (ours)
 & \cellcolor{lightpeach!30}\colorbox{lightpeach!80}{69.40}
 & \cellcolor{lightpeach!30}\colorbox{lightpeach!80}{0.6457}
 & \cellcolor{lightpeach!30}\colorbox{lightpeach!80}{0.7762}
 & \cellcolor{lightpeach!30}\colorbox{lightpeach!80}{5.475} \\
\bottomrule
\end{tabular}%
}%
\end{minipage}%
\hfill
\begin{minipage}[t]{0.4\linewidth}
\centering
\subcaption{Per-axis ablation of ORA.}
\label{tab:ablation_axis}
\resizebox{\linewidth}{!}{%
\setlength\tabcolsep{3pt}
\renewcommand\arraystretch{1.1}
\begin{tabular}{cccccccc}
\toprule
Variant & DP & OR & Tool
 & MUSIQ $\uparrow$ & MANIQA $\uparrow$ & CLIP-IQA+ $\uparrow$ & NIQE $\downarrow$ \\
\midrule
B.1 & $\times$ & \checkmark & \checkmark
 & 65.21 & 0.5735 & 0.6863 & 6.757 \\
B.2 & \checkmark & $\times$ & \checkmark
 & 66.17 & 0.5958 & 0.7099 & 7.032 \\
B.3 & \checkmark & \checkmark & $\times$
 & \colorbox{customblue}{68.65} & \colorbox{lightpeach!80}{0.6504} & \colorbox{customblue}{0.7668} & \colorbox{customblue}{6.235} \\
C & \cellcolor{lightpeach!30}\checkmark & \cellcolor{lightpeach!30}\checkmark & \cellcolor{lightpeach!30}\checkmark
 & \cellcolor{lightpeach!30}\colorbox{lightpeach!80}{69.40}
 & \cellcolor{lightpeach!30}\colorbox{customblue!80}{0.6457}
 & \cellcolor{lightpeach!30}\colorbox{lightpeach!80}{0.7762}
 & \cellcolor{lightpeach!30}\colorbox{lightpeach!80}{5.475} \\
\bottomrule
\end{tabular}%
}%
\end{minipage}%
\end{table}
\vspace{-1.5mm}

\paragraph{Effect of Order-aware Restoration Alignment.}
Tab.~\ref{tab:ablation_ora} validates the effectiveness of ORA.
Variant A, trained only with SFT on $\mathcal{D}_{\text{ORTD}}^{\text{DiTTo}}$, performs the worst because it learns simulator-generated restoration-action-trajectories without alignment to real restoration-experts.
Since this SFT-only agent cannot perform real restoration-expert selection, we parse only the predicted degradation type and select the restoration-expert using IQA-best matching for evaluation.
Variant B improves over Variant A, showing that DPO is useful for reducing the simulator-to-expert gap.
However, Variant C achieves the best overall final restored image-state quality, demonstrating that ORA is more effective than Generic DPO by aligning chosen and rejected responses over decomposed planning axes rather than optimizing a single response-level margin.

\paragraph{Effect of each decomposed planning axis.}
Tab.~\ref{tab:ablation_axis} analyzes the contribution of each decomposed planning axis in ORA.
Removing the DP axis in Variant B.1 substantially degrades final restoration quality, indicating that Degradation Perception-Reasoning is necessary for identifying the involved degradation type set $\mathbf{D}^{\boldsymbol{\delta}}$ before order-aware restoration.
Removing the OR axis in Variant B.2 also leads to clear degradation, confirming that Order-aware Restoration is a core factor in multi-degradation restoration.
Removing the Tool axis in Variant B.3 preserves relatively strong image-quality scores, but still underperforms the full ORA variant on most metrics, suggesting that JSON-based tool-call format mainly stabilizes structured tool use rather than replacing the need for DP and OR alignment.
Variant C, which uses all three decomposed planning axes, achieves the best overall performance, demonstrating that Degradation Perception-Reasoning, Order-aware Restoration, and JSON-based tool-call format provide complementary alignment signals.

\paragraph{Plug-and-Play Restoration-Expert Extensibility.}
\begin{wraptable}{r}{0.55\linewidth}
\vspace{-1em}
\centering
\caption{
Per-stage adaptation cost on $2\times$B200.
Both methods share the same SFT pipeline; the savings come from data generation and alignment.
}
\label{tab:ablation_plugplay}
\resizebox{0.55\linewidth}{!}{%
\setlength\tabcolsep{4pt}
\renewcommand\arraystretch{1.15}
\begin{tabular}{lc|ccc}
\toprule
Stage & JarvisIR~\cite{lin2025jarvisir} & Speedup & DiTTo (ours) \\
\midrule
Data generation & $\sim$460 & $\sim$45$\times$ & $\sim$10 \\
SFT             & $\sim$37  & $1\times$        & $\sim$37 \\
Alignment       & $\sim$410 & $\sim$37$\times$ & $\sim$11 \\
\midrule
\textbf{End-to-end} & $\sim$907 & $\sim$\textbf{15$\times$} & $\sim$60 \\
\bottomrule
\end{tabular}%
}
\vspace{-0.8em}
\end{wraptable}
Our two-stage design decouples agent training from the concrete real-expert set, enabling cheap adaptation when the expert pool changes. We measure \emph{adaptation speed}, i.e., the wall-clock hours to produce a deployable agent for a new expert pool, on $2\times$B200 with training corpus, batch configuration, and hardware matched across both methods. We focus on speed alone, as restoration quality is confounded by model capacity and a full JarvisIR run is prohibitively slow.
Tab.~\ref{tab:ablation_plugplay} decomposes the cost into three stages.
\textbf{(1) Data generation}: JarvisIR enumerates real-expert chains per image with repeated expert execution and IQA scoring; DiTTo replaces this with a single greedy traversal of $\cup$S-IR ($\sim$45$\times$).
\textbf{(2) SFT}: matched by construction.
\textbf{(3) Alignment}: JarvisIR's MRRHF~\cite{lin2025jarvisir} runs online beam-search, expert execution, and multi-metric IQA at every step; DiTTo's ORA is offline DPO over pre-computed ORTD pairs, eliminating all three per-step costs ($\sim$37$\times$).
End-to-end, adapting to a new expert costs $\sim$907h for JarvisIR versus $\sim$60h for DiTTo, a $\sim$15$\times$ reduction that validates the plug-and-play property. Restoration-quality results under both expert-pool and degradation-universe extensions are reported in \textit{Appendix}.~\ref{sec:extensibility}.
\vspace{-1.5mm}
\section{Conclusion}
\label{sec:conclusion}
\vspace{-3mm}
We presented \textbf{DiTTo}, a scalable order-aware image restoration agent framework for multi-degradation All-in-One image restoration.
DiTTo constructs $\mathcal{D}_{\text{ORTD}}^{\text{DiTTo}}$ with the \textbf{DiTTo Simulator}, reducing ORTD pair generation from exhaustive real restoration-expert calls to $\mathcal{O}(N^{\mathbf{D}})$ simulator steps per image.
The DiTTo Agent is then trained by Stage~1 SFT and Stage~2 \textbf{Order-aware Restoration Alignment (ORA)}, which aligns DP, OR, and tool-call axes to reduce the simulator-to-expert gap.
This design enables plug-and-play scalable extensibility by reusing $\cup$S-IR, AiO-IQA, and $W_{\text{DiTTo}}^{\text{SFT}}$ when adding new restoration-experts.
Experiments on MiO-100 validate the effectiveness of DiTTo for state-of-the-art agent-based multi-degradation restoration. A short demo video illustrating the end-to-end inference of DiTTo Agent on real multi-degraded images is included in the supplementary material.

{
    \small
    \bibliographystyle{abbrv}
    \bibliography{references}
}

\newpage
\appendix
\newpage
\section*{Appendix Contents}
\startcontents[appendix]
\printcontents[appendix]{}{1}{\setcounter{tocdepth}{2}}
\newpage

\section{Related Work}
\label{sec:related_work}

\paragraph{All-in-One Image Restoration.}
Early IR methods assume a single known degradation type~\cite{he2010single,nah2017deep,zamir2022restormer}, limiting applicability to real-world scenarios where multiple degradations co-occur.
All-in-one IR addresses this by training a unified model across diverse degradations, using contrastive learning~\cite{li2022airnet}, prompt- or instruction-based conditioning~\cite{potlapalli2023promptir,conde2024instructir}, or iterative re-assessment~\cite{chen2026rar}.
However, these methods learn a direct degraded-to-restored mapping without modeling a restoration-action-trajectory.
As a result, they do not explicitly reason about the restoration-action-ordering, even though degradation removal order significantly affects restoration quality~\cite{10.1145/3532625,lin2024diffbir}.
They also lack a mechanism for sequentially selecting degradation-specific restoration-actions, motivating a shift toward agentic IR.

\paragraph{Training-free Agentic Image Restoration.}
Training-free agentic IR leverages pretrained vision-language models (VLMs) to select and invoke restoration-experts at inference time without training a dedicated order-aware policy~\cite{zhu2025agenticir,jiang2025multi,zuo2026kagent}.
While flexible, these methods must perform order search by applying candidate restoration-actions with real restoration-experts and evaluating the resulting next restored image states at inference time, requiring $\mathcal{O}((N^{\mathbf{D}})^2)$ real restoration-expert calls per image.
Moreover, they often rely on fixed IQA-based heuristic search rather than a learned order-aware restoration policy, yielding only partial order-awareness (Tab.~\ref{tab:comparison}) and suboptimal restoration-action-trajectories when degradation compositions vary across images.
In contrast, training-based agentic IR transfers this $\mathcal{O}((N^{\mathbf{D}})^2)$ search cost from inference to offline ORTD construction, reducing inference to $\mathcal{O}(N^{\mathbf{D}})$ real restoration-expert calls.

\paragraph{Training-based Agentic Image Restoration.}
Training-based agents distill order-aware restoration into a trained policy by supervising the policy with optimal restoration-action image-state pairs along optimal restoration-action-trajectories, reducing inference to $\mathcal{O}(N^{\mathbf{D}})$ real restoration-expert calls.
Recent training-based agentic IR methods scale this paradigm to richer degradations: RestoreAgent~\cite{chen2024restoreagent} and JarvisIR~\cite{lin2025jarvisir} apply SFT on exhaustively generated restoration-action-trajectories; Q-Agent~\cite{zhou2025qagent} uses quality-driven chain-of-thought (CoT) supervision; and TIR-Agent~\cite{zhang2026tiragent} reduces redundant restoration-expert evaluations via trajectory reuse, while SimpleCall~\cite{lu2025simplecall} amortizes quality evaluation via actor-critic optimization.
However, obtaining ground-truth optimal restoration-action image-state pairs requires evaluating candidate restoration-actions at each restored image state, yielding an $\mathcal{O}((N^{\mathbf{D}})^2)$ ORTD pair generation cost that trajectory reuse, e.g., TIR-Agent, only partially amortizes.
Critically, because supervision is tied to a fixed restoration-expert set, adding a new restoration-expert requires regenerating expert-based ORTD pairs and retraining the agent.
DiTTo removes this coupling by using the \textbf{DiTTo Simulator} for $\mathcal{O}(N^{\mathbf{D}})$ ORTD pair generation and a data-efficient DPO-based \textit{Order-aware Restoration Alignment (ORA)} stage for plug-and-play adaptation to new restoration-experts without retraining $\cup$S-IR, AiO-IQA, or the DiTTo-SFT checkpoint.
\label{sec:rw}

\section{Notation}
\label{sec:notation}

We use one primitive per degraded image instance, the degradation
ordering $\boldsymbol{\delta}$, from which all instance-specific
quantities (degradation types, image-states, trajectories, candidate
restoration-actions) are derived by index operations.
Each entry is written so that every symbol on the right-hand side
appears on the left-hand side, and every functional object is given
with its input and output domains.

A running example is fixed throughout.
The universe $\mathbf{D}$ contains $N^{\mathbf{D}}=4$ degradation types,
$\mathbf{D}=\{D_1,D_2,D_3,D_4\}=
\{\text{rain},\text{fog},\text{noise},\text{low-light}\}$,
with $(N^E_{D_1},N^E_{D_2},N^E_{D_3},N^E_{D_4})=(2,3,2,2)$
restoration-experts per type.
A specific instance has degradation-ordering
$\boldsymbol{\delta}=(2,1,4)$ (sequentially apply fog, then rain, then low-light)
and restoration-action-ordering
$\boldsymbol{\rho}^{\boldsymbol{\delta}}=
((D_4,1),(D_2,2),(D_1,1))$
(remove low-light first using its restoration-expert $1$, then fog
using its restoration-expert $2$, then rain using its
restoration-expert $1$).

\textbf{Conventions.} \newline
\textbf{Font.}
Bold (ex) $\mathbf{D}$, $\boldsymbol{\delta}$, $\boldsymbol{\rho}$,
$\boldsymbol{\pi}$, $\hat{\mathbf{q}}$) marks sets, sequences and
vectors;
italic (ex) $D_n$, $i^D$, $i^R$, $n$, $i^E_{D_n}$) marks elements
and indices;
calligraphic (ex) $\mathcal{T}$, $\mathcal{D}$,
$\mathcal{S}$, $\mathcal{P}$, $\mathcal{I}$) marks trajectories,
datasets, the simulator, the permutation set,
and the image-state space.

\textbf{Indices.}
$i^D\in\{0,\ldots,j\}$ is the degradation step (number of
degradation-actions applied so far), counting upward.
$i^R\in\{0,\ldots,j\}$ is the restoration index, counting the number
of degradations \emph{still present} in a restored image-state,
so $i^R$ \emph{decreases} as restoration progresses ($j\to 0$).
$n\in\{1,\ldots,N^{\mathbf{D}}\}$ is the degradation-type index in
$\mathbf{D}$.
$i^E_{D_n}\in\{1,\ldots,N^E_{D_n}\}$ is the restoration-expert index
inside the restoration-expert pool of degradation type $D_n$
its range depends on the paired type, so we always write
$i^E_{D_n}$ rather than a bare $i^E$.
The indices $i^D$ and $i^R$ themselves are instance-independent;
only the values they index are instance-dependent.

\textbf{Side superscripts.}
$D$ marks degradation, $R$ marks restoration, $E$ marks
restoration-experts.
The superscript $\boldsymbol{\delta}$ on any symbol means
``the value this symbol takes in the instance with
degradation-ordering $\boldsymbol{\delta}$'';
a trailing $*$ marks optimality;
tildes ($\widetilde{\,\cdot\,}$) mark restored image-states.

\textbf{Order-awareness.}
Both $\boldsymbol{\delta}$ and $\boldsymbol{\rho}^{\boldsymbol{\delta}}$
are ordered tuples, written with parentheses to distinguish them
from sets;
permuting their entries yields a different instance or a different
degradation-action- or restoration-action-trajectory, even when the
underlying type sets coincide.

\textbf{Functional notation.}
Let $\mathcal{I}=\mathbb{R}^{H\times W\times C}$ denote the space of
image-states, where $H,W,C\in\mathbb{Z}_{>0}$ are the height, width
and number of channels of the image, fixed throughout an instance.
Functions are declared with their input and output domains, e.g.,
$A^D_{D_n}(\cdot):\mathcal{I}\to\mathcal{I}$.
Application fills the slot with a specific element $I\in\mathcal{I}$,
e.g., $A^D_{D_n}(I)$.

\textbf{Shorthand.}
When $\boldsymbol{\delta}$ is unambiguous, its superscript is
dropped.

\textbf{Degradation universe and restoration-experts.}
\begin{description}
  \item[$\mathbf{D}=\{D_1,\ldots,D_{N^{\mathbf{D}}}\}$:]
  The predefined universe of degradation types, fixed across all
  images, with cardinality $|\mathbf{D}|=N^{\mathbf{D}}$.
  E.g., $\mathbf{D}=\{\text{rain},\text{fog},\text{noise},
  \text{low-light}\}$ with $N^{\mathbf{D}}=4$.

  \item[$D_n,\ n\in\{1,\ldots,N^{\mathbf{D}}\}$:]
  The $n$-th degradation type itself; the symbol denotes the type
  and the subscript $n$ is its index within $\mathbf{D}$.
  E.g., $D_1=\text{rain}$, $D_2=\text{fog}$, $D_3=\text{noise}$,
  $D_4=\text{low-light}$.

  \item[$A^D_{D_n}(\cdot):\mathbb{R}^{H\times W\times C}\to
  \mathbb{R}^{H\times W\times C}$:]
  The degradation-action that applies degradation type $D_n$ to its
  input image-state.
  E.g., $A^D_{D_1}$ adds rain; $A^D_{D_4}$ darkens the image.

  \item[$A^R_{(D_n,i^E_{D_n})}(\cdot):
  \mathbb{R}^{H\times W\times C}\to\mathbb{R}^{H\times W\times C}$:]
  The restoration-action realised by a specific restoration-expert,
  identified jointly by the degradation type $D_n$ it specialises
  in and the restoration-expert index
  $i^E_{D_n}\in\{1,\ldots,N^E_{D_n}\}$ within that type's pool.
  Applied to an image-state, it removes degradation $D_n$ and
  \textit{leaves any other degradations present in the input intact
  (single-degradation restoration)}.
  E.g., $A^R_{(D_1,1)}$ and $A^R_{(D_1,2)}$ are two specific
  de-rainers (e.g., \texttt{Restormer}, \texttt{IDT});
  $A^R_{(D_2,3)}$ is the third de-fogger in the pool of $D_2$.

  \item[$\mathbf{A}^R_{D_n}=\{A^R_{(D_n,i^E_{D_n})}(\cdot):
  i^E_{D_n}\in\{1,\ldots,N^E_{D_n}\}\}$:]
  The restoration-action pool for degradation type $D_n$:
  the set of all restoration-actions indexed by $i^E_{D_n}$ for
  that type, with type-dependent cardinality
  $|\mathbf{A}^R_{D_n}|=N^E_{D_n}$.
  E.g., $\mathbf{A}^R_{D_1}$ contains $N^E_{D_1}=2$ de-raining
  restoration-actions;
  $\mathbf{A}^R_{D_2}$ contains $N^E_{D_2}=3$ de-fogging
  restoration-actions.
\end{description}

\textbf{Degradation process.}
\begin{description}
  \item[$I_{\text{clean}}\in\mathbb{R}^{H\times W\times C}$:]
  The universe-level clean reference image-state, shared across
  instances and serving as the start of every
  degradation-action-trajectory.

  \item[$\boldsymbol{\delta}=(\delta_{i^D})_{i^D=1}^{j},\ j\geq 1$:]
  The degradation-ordering: an ordered tuple of $j$ distinct entries
  $\delta_{i^D}\in\{1,\ldots,N^{\mathbf{D}}\}$ indexed by
  $i^D\in\{1,\ldots,j\}$, each entry being the index of a
  degradation type in $\mathbf{D}$.
  The length $j$ of the tuple is the number of degradations applied
  in this instance.
  E.g., $\boldsymbol{\delta}=(2,1,4)$ encodes fog, then rain, then
  low-light, with $j=3$. 

  \item[$\delta_{i^D},\ i^D\in\{1,\ldots,j\}$:]
  The $i^D$-th entry of $\boldsymbol{\delta}$, an integer in
  $\{1,\ldots,N^{\mathbf{D}}\}$.
  The corresponding degradation type is $D_{\delta_{i^D}}$ and the
  corresponding degradation-action is
  $A^D_{D_{\delta_{i^D}}}(\cdot)$.
  E.g., for $\boldsymbol{\delta}=(2,1,4)$, $\delta_2=1$, so
  $D_{\delta_2}=D_1=\text{rain}$.

  \item[$\mathbf{D}^{\boldsymbol{\delta}}=
  \{D_{\delta_{i^D}}:i^D\in\{1,\ldots,j\}\}\subseteq\mathbf{D}$:]
  The set of degradation types involved in this instance,
  obtained by collecting the types pointed to by the entries of
  $\boldsymbol{\delta}$, with cardinality
  $|\mathbf{D}^{\boldsymbol{\delta}}|=j$ (entries are distinct).
  E.g., for $\boldsymbol{\delta}=(2,1,4)$,
  $\mathbf{D}^{\boldsymbol{\delta}}=\{D_1,D_2,D_4\}$.

  \item[$I^{\boldsymbol{\delta}}_{i^D}\in
  \mathbb{R}^{H\times W\times C},\ i^D\in\{0,\ldots,j\}$:]
  The degraded image-state after $i^D$ degradation-action steps in
  this instance:
  $I^{\boldsymbol{\delta}}_0=I_{\text{clean}}$ and
  $I^{\boldsymbol{\delta}}_{i^D}=A^D_{D_{\delta_{i^D}}}
  (I^{\boldsymbol{\delta}}_{i^D-1})$ for $i^D\geq 1$.
  At inference, only the final degraded image-state
  $I^{\boldsymbol{\delta}}_j$ (the observed multi-degradation input)
  is given; $\boldsymbol{\delta}$ itself is hidden.
  E.g., for $\boldsymbol{\delta}=(2,1,4)$,
  $I^{\boldsymbol{\delta}}_1$ is foggy,
  $I^{\boldsymbol{\delta}}_2$ adds rain on top, and
  $I^{\boldsymbol{\delta}}_3$ further darkens it.

  \item[$\mathcal{T}^{\boldsymbol{\delta}}=
  \{(I^{\boldsymbol{\delta}}_{i^D-1},
  A^D_{D_{\delta_{i^D}}}(\cdot),
  I^{\boldsymbol{\delta}}_{i^D})\}_{i^D=1}^{j}$:]
  The degradation-action-trajectory of this instance: the sequence
  of (degraded image-state, degradation-action, next degraded image
  state) triples for $i^D=1,\ldots,j$ in application order.
\end{description}

\begin{figure}[h]
    \centering
    \includegraphics[width=\linewidth]{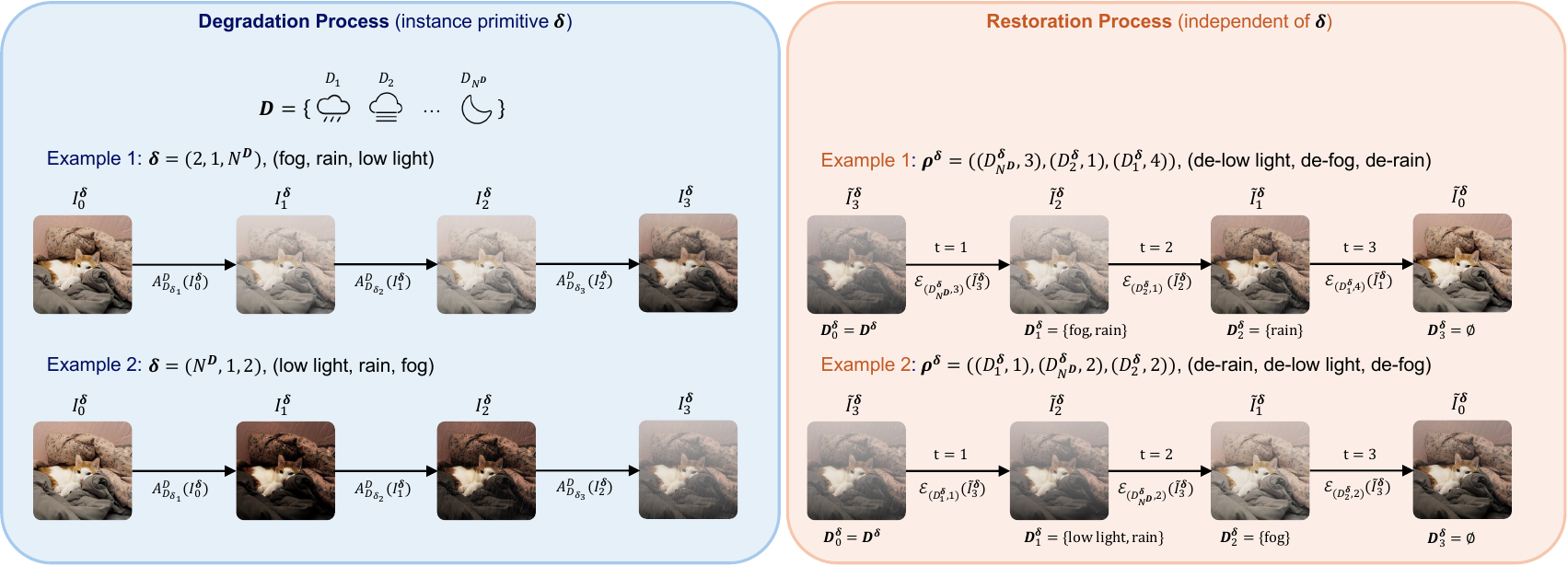}
    \caption{%
      Degradation process (left) and restoration process (right) for
      two instances sharing the type set
      $\mathbf{D}^{\boldsymbol{\delta}}=\{D_1,D_2,D_{N^{\mathbf{D}}}\}$
      but using different degradation-orderings $\boldsymbol{\delta}$,
      so they are distinct instances.
      The degradation side is indexed by $i^D\in\{0,\ldots,j\}$
      (counts degradation-actions applied);
      the restoration side is indexed by $i^R\in\{0,\ldots,j\}$
      (counts degradations still present), so the
      restoration-action-trajectory runs from
      $\widetilde{I}^{\boldsymbol{\delta}}_j:=I^{\boldsymbol{\delta}}_j$
      down to the restored clean image-state
      $\widetilde{I}^{\boldsymbol{\delta}}_0$.
      The restoration-action-ordering
      $\boldsymbol{\rho}^{\boldsymbol{\delta}}$
      is independent of $\boldsymbol{\delta}$:
      the two examples share the same type set but choose different
      removal orders and different restoration-experts.
      Below each restored image-state
      $\widetilde{I}^{\boldsymbol{\delta}}_{i^R}$ we show the
      remaining-type set $\mathbf{D}^{\boldsymbol{\delta}}_{i^R}$.%
    }
    \label{fig:degradation_process}
\end{figure}

\textbf{Restoration process.} \\
The restoration-action-trajectory starts from the observed
multi-degraded image-state
$\widetilde{I}^{\boldsymbol{\delta}}_j:=I^{\boldsymbol{\delta}}_j$
and ends at the restored clean image-state
$\widetilde{I}^{\boldsymbol{\delta}}_0$;
the index $i^R$ counts degradations \emph{still present}, decreasing
by one per restoration-action step
($j\!\to\!j{-}1\!\to\!\cdots\!\to\!0$).
The restoration-action-ordering is \emph{not constrained by
$\boldsymbol{\delta}$}:
any permutation of $\mathbf{D}^{\boldsymbol{\delta}}$, paired with
any restoration-expert per step, is allowed. 

\begin{description}
  \item[$D^{\boldsymbol{\delta}}_{i^R}\in
  \mathbf{D}^{\boldsymbol{\delta}}$,
  $i^R\in\{1,\ldots,j\}$:]
  The degradation type chosen for removal at restoration-action
  index $i^R$, i.e., at the transition
  $\widetilde{I}^{\boldsymbol{\delta}}_{i^R}\!
  \to\!\widetilde{I}^{\boldsymbol{\delta}}_{i^R-1}$.
  E.g., for $\boldsymbol{\rho}^{\boldsymbol{\delta}}=
  ((D_4,1),(D_2,2),(D_1,1))$,
  $D^{\boldsymbol{\delta}}_3=D_4$, $D^{\boldsymbol{\delta}}_2=D_2$,
  $D^{\boldsymbol{\delta}}_1=D_1$.

  \item[$\rho^{\boldsymbol{\delta}}_{i^R}=
  (D^{\boldsymbol{\delta}}_{i^R},
  i^E_{D^{\boldsymbol{\delta}}_{i^R}}),\
  i^R\in\{1,\ldots,j\}$:]
  The restoration-action step at index $i^R$:
  the chosen degradation type $D^{\boldsymbol{\delta}}_{i^R}$ paired
  with a restoration-expert index
  $i^E_{D^{\boldsymbol{\delta}}_{i^R}}$ within that type's pool;
  the corresponding restoration-action is
  $A^R_{\rho^{\boldsymbol{\delta}}_{i^R}}(\cdot)$.
  E.g., $\rho^{\boldsymbol{\delta}}_3=(D_4,1)$ removes low-light
  using the first restoration-expert in the pool of $D_4$. 

  \item[$\boldsymbol{\rho}^{\boldsymbol{\delta}}=
  (\rho^{\boldsymbol{\delta}}_{i^R})_{i^R=j}^{1}$:]
  The restoration-action-ordering for this instance: an ordered
  tuple indexed by $i^R$ running from $j$ down to $1$, so that
  $\rho^{\boldsymbol{\delta}}_j$ is applied first and
  $\rho^{\boldsymbol{\delta}}_1$ last.
  Valid if and only if the chosen types cover the involved type set,
  i.e.,
  $\{D^{\boldsymbol{\delta}}_{i^R}:i^R\in\{1,\ldots,j\}\}
  =\mathbf{D}^{\boldsymbol{\delta}}$.
  E.g., $\boldsymbol{\rho}^{\boldsymbol{\delta}}=
  ((D_4,1),(D_2,2),(D_1,1))$ for the running example.

  \item[$\widetilde{I}^{\boldsymbol{\delta}}_{i^R}\in
  \mathbb{R}^{H\times W\times C},\ i^R\in\{0,\ldots,j\}$:]
  The restored image-state with $i^R$ degradations still
  remaining:
  $\widetilde{I}^{\boldsymbol{\delta}}_j:=I^{\boldsymbol{\delta}}_j$
  and $\widetilde{I}^{\boldsymbol{\delta}}_{i^R-1}=
  A^R_{\rho^{\boldsymbol{\delta}}_{i^R}}
  (\widetilde{I}^{\boldsymbol{\delta}}_{i^R})$
  for $i^R\geq 1$.
  E.g., for the running example,
  $\widetilde{I}^{\boldsymbol{\delta}}_3$ is the observed input
  (rain + fog + low-light),
  $\widetilde{I}^{\boldsymbol{\delta}}_2$ has low-light removed,
  $\widetilde{I}^{\boldsymbol{\delta}}_1$ further has fog removed,
  and $\widetilde{I}^{\boldsymbol{\delta}}_0$ is the final
  restored clean image-state.

  \item[$\mathbf{D}^{\boldsymbol{\delta}}_{i^R}=
  \mathbf{D}^{\boldsymbol{\delta}}
  \setminus\{D^{\boldsymbol{\delta}}_{i^{R'}}:i^{R'}>i^R\},\
  i^R\in\{0,\ldots,j\}$:]
  The remaining-degradation type set at restoration index $i^R$, written in
  closed form as the involved type set minus the types already
  removed at later (higher-$i^R$) steps.
  Boundaries: $\mathbf{D}^{\boldsymbol{\delta}}_j=
  \mathbf{D}^{\boldsymbol{\delta}}$ at the start and
  $\mathbf{D}^{\boldsymbol{\delta}}_0=\varnothing$ at the end.
  E.g., for $\boldsymbol{\rho}^{\boldsymbol{\delta}}=
  ((D_4,1),(D_2,2),(D_1,1))$:
  $\mathbf{D}^{\boldsymbol{\delta}}_3=\{D_1,D_2,D_4\}$,
  $\mathbf{D}^{\boldsymbol{\delta}}_2=\{D_1,D_2\}$,
  $\mathbf{D}^{\boldsymbol{\delta}}_1=\{D_1\}$,
  $\mathbf{D}^{\boldsymbol{\delta}}_0=\varnothing$.

  \item[$\mathbf{A}^{\boldsymbol{\delta}}_{i^R}=
  \{(D,i^E_D):D\in\mathbf{D}^{\boldsymbol{\delta}}_{i^R},\,
  i^E_D\in\{1,\ldots,N^E_D\}\},\
  i^R\in\{1,\ldots,j\}$:]
  The candidate restoration-action set at index $i^R$:
  every (remaining degradation type, restoration-expert index) pair
  available for selection, with cardinality
  $|\mathbf{A}^{\boldsymbol{\delta}}_{i^R}|=
  \sum_{D\in\mathbf{D}^{\boldsymbol{\delta}}_{i^R}}N^E_D$ summing
  the pool sizes of the still-remaining types.
  $\rho^{\boldsymbol{\delta}}_{i^R}$ is selected from
  $\mathbf{A}^{\boldsymbol{\delta}}_{i^R}$.
  E.g., at $i^R=3$ in the running example,
  $|\mathbf{A}^{\boldsymbol{\delta}}_3|=N^E_{D_1}+N^E_{D_2}+N^E_{D_4}
  =2+3+2=7$;
  after the first restoration-action removes $D_4$, at $i^R=2$ we
  have $|\mathbf{A}^{\boldsymbol{\delta}}_2|=
  N^E_{D_1}+N^E_{D_2}=2+3=5$. 

  \item[$\mathcal{P}^{\boldsymbol{\delta}}$:]
  The set of all valid restoration-action-orderings on
  $\mathbf{D}^{\boldsymbol{\delta}}$:
  every permutation of the involved degradation types, paired
  step-by-step with any valid restoration-expert for that type,
  with cardinality
  $|\mathcal{P}^{\boldsymbol{\delta}}|=
  j!\,\prod_{D\in\mathbf{D}^{\boldsymbol{\delta}}}N^E_D$.
  E.g., for the running example,
  $|\mathcal{P}^{\boldsymbol{\delta}}|
  =3!\cdot N^E_{D_1}\cdot N^E_{D_2}\cdot N^E_{D_4}
  =6\cdot 2\cdot 3\cdot 2=72$. 

  \item[$\mathcal{T}^{\boldsymbol{\delta},
  \boldsymbol{\rho}^{\boldsymbol{\delta}}}=
  \{(\widetilde{I}^{\boldsymbol{\delta}}_{i^R},
  A^R_{\rho^{\boldsymbol{\delta}}_{i^R}}(\cdot),
  \widetilde{I}^{\boldsymbol{\delta}}_{i^R-1})\}_{i^R=j}^{1}$:]
  The restoration-action-trajectory induced by
  $\boldsymbol{\rho}^{\boldsymbol{\delta}}$: the sequence of
  (restored image-state, restoration-action, next restored image
  state) triples in application order ($i^R=j$ first, $i^R=1$ last).
\end{description}

\textbf{Optimal restoration.}
\begin{description}
  \item[$Q(\cdot):\mathbb{R}^{H\times W\times C}\to\mathbb{R}$:]
  An image-quality score (higher is better) on a given image,
  instantiated by an IQA metric (or a fixed scalar combination of
  several IQA metrics). 

  \item[$\boldsymbol{\pi}^{\boldsymbol{\delta}}=
  (\pi^{\boldsymbol{\delta}}_{i^R})_{i^R=j}^{1}$:]
  The optimal restoration-action-ordering for this instance:
  the element of $\mathcal{P}^{\boldsymbol{\delta}}$ whose induced
  restoration-action-trajectory yields the highest quality score $Q$
  on the final restored image-state.
  $\boldsymbol{\pi}^{\boldsymbol{\delta}}$ has the same loop format
  as $\boldsymbol{\rho}^{\boldsymbol{\delta}}$;
  each entry is denoted as $\pi^{\boldsymbol{\delta}}_{i^R}=
  (D^{\boldsymbol{\delta}}_{i^R},
  i^{E,*}_{D^{\boldsymbol{\delta}}_{i^R}})$,
  where the asterisk marks joint optimality of the chosen type and
  restoration-expert.
  When the maximiser is non-unique, ties are broken by a fixed
  deterministic rule.

  \item[$\mathcal{T}^{\boldsymbol{\delta},*}$:]
  The optimal restoration-action-trajectory: the trajectory induced
  by $\boldsymbol{\pi}^{\boldsymbol{\delta}}$ in place of a generic
  $\boldsymbol{\rho}$, i.e.,
  $\mathcal{T}^{\boldsymbol{\delta},*}=
  \mathcal{T}^{\boldsymbol{\delta},
  \boldsymbol{\pi}^{\boldsymbol{\delta}}}$.
\end{description}

\textbf{Agents.}
\begin{description}
  \item[$\mathit{Agent}_{DP}(\cdot):
  \mathbb{R}^{H\times W\times C}\to 2^{\mathbf{D}}$:]
  The degradation-perception agent.
  Given the observed multi-degraded image-state
  $I^{\boldsymbol{\delta}}_j$, it returns a subset of $\mathbf{D}$:
  the involved degradation types
  $\mathbf{D}^{\boldsymbol{\delta}}=
  \mathit{Agent}_{DP}(I^{\boldsymbol{\delta}}_j)$
  (and hence $j=|\mathbf{D}^{\boldsymbol{\delta}}|$);
  here $2^{\mathbf{D}}$ denotes the power set of $\mathbf{D}$ (the
  set of all subsets of $\mathbf{D}$).
  Assumed perfect throughout the analysis.
  E.g., for the running example, given $I^{\boldsymbol{\delta}}_3$,
  $\mathit{Agent}_{DP}$ returns $\{D_1,D_2,D_4\}$.

  \item[$\mathit{Agent}_{OR}(\cdot,\cdot):
  \mathbb{R}^{H\times W\times C}\times 2^{\mathbf{D}}\to
  \bigcup_{\boldsymbol{\delta}}\mathcal{P}^{\boldsymbol{\delta}}$:]
  The optimal-restoration agent.
  Given $(I^{\boldsymbol{\delta}}_j,\mathbf{D}^{\boldsymbol{\delta}})$,
  it selects a restoration-action-ordering in
  $\mathcal{P}^{\boldsymbol{\delta}}$, used as
  $\boldsymbol{\pi}^{\boldsymbol{\delta}}=
  \mathit{Agent}_{OR}(I^{\boldsymbol{\delta}}_j,
  \mathbf{D}^{\boldsymbol{\delta}})$ at inference.

  \item[$\mathit{Agent}_{DP+OR}(\cdot):
  \mathbb{R}^{H\times W\times C}\to
  \bigcup_{\boldsymbol{\delta}}\mathcal{P}^{\boldsymbol{\delta}}$:]
  The DiTTo agent: the composition that runs
  $\mathit{Agent}_{DP}$ on the observed input and then
  $\mathit{Agent}_{OR}$ on its result, defined by
  \[
    \mathit{Agent}_{DP+OR}(I)=
    \mathit{Agent}_{OR}\big(I,\mathit{Agent}_{DP}(I)\big).
  \]
\end{description}

\textbf{ORTD and DiTTo simulator.}
\begin{description}
  \item[$\mathcal{D}_{\text{ORTD}}=\bigcup_{\boldsymbol{\delta}}
  \{(\widetilde{I}^{\boldsymbol{\delta}}_{i^R},
  A^R_{\pi^{\boldsymbol{\delta}}_{i^R}}(\cdot))
  \}_{i^R=j}^{1}$:]
  The Optimal Restoration-action Trajectory Dataset:
  the union, over training instances $\boldsymbol{\delta}$, of all
  (restored image-state, restoration-action) pairs along the
  optimal restoration-action-trajectory
  $\mathcal{T}^{\boldsymbol{\delta},*}$.
  Two construction variants:
  $\mathcal{D}_{\text{ORTD}}^{\text{Expert}}$ uses real
  restoration-experts ($\mathcal{O}((N^{\mathbf{D}})^2)$
  restoration-expert calls per image);
  $\mathcal{D}_{\text{ORTD}}^{\text{DiTTo}}$ uses the simulator
  $\mathcal{S}_\theta$ ($\mathcal{O}(N^{\mathbf{D}})$ rollouts per
  image).

  \item[$\mathcal{S}_\theta(\cdot,\cdot):
  \mathbb{R}^{H\times W\times C}\times
  (\mathbf{D}\times\mathbb{Z}_{>0})\to
  \mathbb{R}^{H\times W\times C}$:]
  The single-step restoration simulator with parameters $\theta$.
  Input: a current restored image-state and a restoration-action
  specified by a (degradation type, restoration-expert index) pair;
  output: the restored image-state after that single
  restoration-action, with the specified degradation removed and
  the rest unchanged.
  Used in place of an actual restoration-expert during
  $\mathcal{D}_{\text{ORTD}}^{\text{DiTTo}}$ construction.
  E.g., $\mathcal{S}_\theta(\widetilde{I}^{\boldsymbol{\delta}}_3,
  (D_4,1))=\widetilde{I}^{\boldsymbol{\delta}}_2$. 

  \item[$f_{\psi}(\cdot,\cdot):
  \mathbb{R}^{H\times W\times C}\times
  \{\mathcal{T}^{\boldsymbol{\delta}}\}_{\boldsymbol{\delta}}
  \to {[0,1]}^{|\mathbf{A}^{\boldsymbol{\delta}}_{i^R}|}$:]
  The all-in-one IQA-style scoring model with parameters $\psi$.
  Input: a current restored image-state and the degradation-action
  trajectory $\mathcal{T}^{\boldsymbol{\delta}}$ of the instance;
  output: a score vector in
  $[0,1]^{|\mathbf{A}^{\boldsymbol{\delta}}_{i^R}|}$ predicting the
  quality of all candidate restoration-actions at the current
  restoration indeX.

  \item[$\hat{\mathbf{q}}^{\boldsymbol{\delta}}_{i^R}=
  f_{\psi}(\widetilde{I}^{\boldsymbol{\delta}}_{i^R},
  \mathcal{T}^{\boldsymbol{\delta}})
  \in {[0,1]}^{|\mathbf{A}^{\boldsymbol{\delta}}_{i^R}|}$:]
  The score vector at restoration index $i^R$.
  Each entry of $\hat{\mathbf{q}}^{\boldsymbol{\delta}}_{i^R}$ is
  aligned one-to-one with a candidate restoration-action
  $a\in\mathbf{A}^{\boldsymbol{\delta}}_{i^R}$ under a fixed
  enumeration, where
  $a=(D,i^E_D)$ is a (remaining degradation type,
  restoration-expert index) pair;
  the entry's value is the predicted quality of applying the
  corresponding restoration-action to the current restored image-state
  $\widetilde{I}^{\boldsymbol{\delta}}_{i^R}$, conditioned on
  the degradation-action-trajectory $\mathcal{T}^{\boldsymbol{\delta}}$
  that generated the observed multi-degraded instance.
  The vector therefore has one scalar entry per candidate, with
  total length $|\mathbf{A}^{\boldsymbol{\delta}}_{i^R}|$, and its
  entries are aligned one-to-one with the candidates in
  $\mathbf{A}^{\boldsymbol{\delta}}_{i^R}$ under a fixed enumeration.
  At inference, $\rho^{\boldsymbol{\delta}}_{i^R}$ is the candidate
  whose aligned entry is the highest in
  $\hat{\mathbf{q}}^{\boldsymbol{\delta}}_{i^R}$
  (ties broken deterministically);
  no masking is needed because
  $\mathbf{A}^{\boldsymbol{\delta}}_{i^R}$ already restricts
  candidates to the remaining types.
\end{description}

\definecolor{usircolor}{RGB}{30,90,160}    
\definecolor{aiocolor}{RGB}{180,70,30}     
\newcommand{\usir}[1]{\textcolor{usircolor}{#1}}
\newcommand{\aio}[1]{\textcolor{aiocolor}{#1}}

\section{Algorithm}
\label{algorithms}

This section provides the pseudo-code for the three core procedures
that constitute the DiTTo Simulator pipeline introduced in
Sec.~\ref{sec:method} of the main paper:
training $\cup$S-IR (Algorithm~\ref{alg:train_usir}), training
AiO-IQA on top of frozen $\cup$S-IR
(Algorithm~\ref{alg:train_aio_iqa}), and constructing
$\mathcal{D}_{\text{ORTD}}^{\text{DiTTo}}$ from the two trained
modules (Algorithm~\ref{alg:construct_ditto_ortd}).
All three procedures share the notation in
Sec.~\ref{sec:notation}: a degradation-ordering
$\boldsymbol{\delta}=(\delta_{i^D})_{i^D=1}^{j}$ instantiates a
degradation-action-trajectory $\mathcal{T}^{\boldsymbol{\delta}}$
that produces the observed multi-degraded image-state
$I^{\boldsymbol{\delta}}_j$, and the restoration index
$i^R\!\in\!\{0,\ldots,j\}$ counts the degradations still present in
the current restored image-state $\widetilde{I}^{\boldsymbol{\delta}}_{i^R}$.

\subsection{Training $\cup$S-IR}

Algorithm~\ref{alg:train_usir} trains the single-degradation
restoration simulator \usir{$\mathcal{S}_{\theta}$} to approximate the
effect of an arbitrary candidate restoration-action identifier
$\rho\in\mathbf{A}^{\boldsymbol{\delta}}_{i^R}$ on the current
restored image-state $\widetilde{I}^{\boldsymbol{\delta}}_{i^R}$.
At every iteration we sample a clean image-state $I_{\text{clean}}$
and a degradation-ordering $\boldsymbol{\delta}$ to construct
$\mathcal{T}^{\boldsymbol{\delta}}$ and the observed multi-degraded
image-state $I^{\boldsymbol{\delta}}_j$.
We then sample one restoration index $i^R$ and one candidate
identifier $\rho\in\mathbf{A}^{\boldsymbol{\delta}}_{i^R}$, and build
the supervision target $\widetilde{I}^{\boldsymbol{\delta}}_{i^R-1}$
by removing only the degradation type $D$ specified by $\rho$ while
\emph{preserving the other remaining degradations} in
$\mathbf{D}^{\boldsymbol{\delta}}_{i^R}\setminus\{D\}$.
The model output
\[
\widehat{I}^{\boldsymbol{\delta}}_{i^R-1,\rho}
=
\usir{\mathcal{S}_{\theta}}\!\left(
\widetilde{I}^{\boldsymbol{\delta}}_{i^R},\rho\right)
\]
is the predicted next restored image-state in which $D$ is removed
while the remaining degradations are retained, and
\usir{$\mathcal{S}_{\theta}$} is updated by the SD3-style
flow-matching objective~\citep{esser2024scaling} (Sec.~\ref{sec:usir_loss})
between this prediction and $\widetilde{I}^{\boldsymbol{\delta}}_{i^R-1}$.
The restoration index $i^R$ is uniformly sampled across
$\{1,\ldots,j\}$ so that \usir{$\mathcal{S}_{\theta}$} sees every stage
of the restoration-action-trajectory during training.

\begin{algorithm}[h]
\caption{Training \usir{$\cup$S-IR ($\mathcal{S}_{\theta}$)}}
\label{alg:train_usir}
\begin{algorithmic}[1]
\Require Clean image-states $I_{\text{clean}}$; degradation universe $\mathbf{D}$; degradation-actions $\{A^D_{D_n}(\cdot)\}_{D_n\in\mathbf{D}}$; \usir{single-degradation restoration simulator $\mathcal{S}_{\theta}$}; total iterations $T$.
\Ensure Trained \usir{$\mathcal{S}_{\theta}$}.

\State Initialize \usir{$\mathcal{S}_{\theta}$}.
\For{$t=1,\ldots,T$}
    \State Sample a clean image-state $I_{\text{clean}}$.
    \State Sample a degradation-ordering $\boldsymbol{\delta}=(\delta_{i^D})_{i^D=1}^{j}$.
    \State Construct the degradation-action-trajectory $\mathcal{T}^{\boldsymbol{\delta}}$ by sequentially applying $A^D_{D_{\delta_{i^D}}}(\cdot)$ to $I_{\text{clean}}$.
    \State Obtain the observed multi-degraded image-state $I^{\boldsymbol{\delta}}_j$ and set $\widetilde{I}^{\boldsymbol{\delta}}_j:=I^{\boldsymbol{\delta}}_j$.
    \State Sample a restoration index $i^R\in\{1,\ldots,j\}$.
    \State Construct the remaining-degradation type set $\mathbf{D}^{\boldsymbol{\delta}}_{i^R}$ and the candidate restoration-action set $\mathbf{A}^{\boldsymbol{\delta}}_{i^R}$.
    \State Sample a candidate restoration-action identifier $\rho=(D,i^E_D)\in\mathbf{A}^{\boldsymbol{\delta}}_{i^R}$.
    \State Construct the supervision target $\widetilde{I}^{\boldsymbol{\delta}}_{i^R-1}$ by removing only the degradation type $D$ while preserving the remaining degradations in $\mathbf{D}^{\boldsymbol{\delta}}_{i^R}\setminus\{D\}$.
    \State Predict the next restored image-state, in which $D$ is removed while the remaining degradations are retained:
    \[
        \widehat{I}^{\boldsymbol{\delta}}_{i^R-1,\rho}
        =
        \usir{\mathcal{S}_{\theta}}\!\left(
        \widetilde{I}^{\boldsymbol{\delta}}_{i^R},\rho\right).
    \]
    \State Sample $\sigma\sim\text{LogitNormal}$ and noise $\boldsymbol{\epsilon}\sim\mathcal{N}(0,\mathbf{I})$, and form $\mathbf{z}_t=(1-\sigma)\,\mathbf{z}_{\text{target}}+\sigma\,\boldsymbol{\epsilon}$.
    \State Update \usir{$\mathcal{S}_{\theta}$} with the SD3-style flow-matching objective~\citep{esser2024scaling}:
    \[
        \mathcal{L}_{\text{FM}}(\usir{\mathcal{S}_{\theta}})
        =
        \mathbb{E}_{\sigma,\boldsymbol{\epsilon}}\!\left[
        w(\sigma)\,
        \sqrt{(\widehat{\mathbf{z}}_0-\mathbf{z}_{\text{target}})^2+\epsilon^2}
        \right],
    \]
    where $\widehat{\mathbf{z}}_0=\mathbf{z}_t-\sigma\,\mathbf{v}_{\usir{\theta}}$ and $\mathbf{z}_{\text{target}}$ is the VAE latent of $\widetilde{I}^{\boldsymbol{\delta}}_{i^R-1}$.
\EndFor
\State \Return \usir{$\mathcal{S}_{\theta}$}.
\end{algorithmic}
\end{algorithm}

\subsection{Training AiO-IQA with Frozen $\cup$S-IR}

Algorithm~\ref{alg:train_aio_iqa} trains the per-action scoring
model \aio{$f_{\psi}$} on top of a frozen \usir{$\mathcal{S}_{\theta}$}.
The key motivation is that supervising \aio{$f_{\psi}$} requires
per-action quality labels at every restoration index, and obtaining
such labels with real restoration-experts would re-introduce the
$\mathcal{O}((N^{\mathbf{D}})^2)$ cost discussed in
Sec.~\ref{sec:method}.
Algorithm~\ref{alg:train_aio_iqa} avoids this by using
\usir{$\mathcal{S}_{\theta}$} as a cheap surrogate: for every
candidate identifier $\rho\in\mathbf{A}^{\boldsymbol{\delta}}_{i^R}$
at the sampled $i^R$, it produces a candidate next restored
image-state via \usir{$\mathcal{S}_{\theta}$} and labels it with the
image-quality score $Q(\cdot)$, defined as a fixed scalar combination
of full-reference (PSNR, SSIM~\citep{wang2004ssim}) and no-reference
(MUSIQ~\citep{ke2021musiq}, MANIQA~\citep{yang2022maniqa},
CLIP-IQA~\citep{wang2023exploring}, NIQE~\citep{mittal2012making})
IQA metrics (Sec.~\ref{sec:aioiqa_ensemble}).
The model output
\[
\hat{\mathbf{q}}^{\boldsymbol{\delta}}_{i^R}
=
\aio{f_{\psi}}\!\left(
\widetilde{I}^{\boldsymbol{\delta}}_{i^R},
\mathcal{T}^{\boldsymbol{\delta}}\right)
\in [0,1]^{|\mathbf{A}^{\boldsymbol{\delta}}_{i^R}|}
\]
is then trained against these labels using the pairwise ranking
objective $\mathcal{L}_{\text{rank}}$ in Sec.~\ref{sec:aioiqa_loss},
which matches both the rank ordering and the relative gaps between
candidates rather than absolute IQA values.
\usir{$\mathcal{S}_{\theta}$} is kept frozen throughout, so it acts
purely as a generator of supervision signal for \aio{$f_{\psi}$}.

\begin{algorithm}[h]
\caption{Training \aio{AiO-IQA ($f_{\psi}$)} with frozen \usir{$\mathcal{S}_{\theta}$}}
\label{alg:train_aio_iqa}
\begin{algorithmic}[1]
\Require Trained \usir{$\mathcal{S}_{\theta}$}; clean image-states $I_{\text{clean}}$; degradation universe $\mathbf{D}$; image-quality score $Q(\cdot)$ (Sec.~\ref{sec:aioiqa_ensemble}); \aio{AiO-IQA model $f_{\psi}$}; total iterations $T$.
\Ensure Trained \aio{$f_{\psi}$}.

\State Freeze \usir{$\mathcal{S}_{\theta}$}; initialize \aio{$f_{\psi}$}.
\For{$t=1,\ldots,T$}
    \State Sample a clean image-state $I_{\text{clean}}$.
    \State Sample a degradation-ordering $\boldsymbol{\delta}=(\delta_{i^D})_{i^D=1}^{j}$.
    \State Construct $\mathcal{T}^{\boldsymbol{\delta}}$ and obtain $I^{\boldsymbol{\delta}}_j$; set $\widetilde{I}^{\boldsymbol{\delta}}_j:=I^{\boldsymbol{\delta}}_j$.
    \State Sample a restoration index $i^R\in\{1,\ldots,j\}$.
    \State Construct $\mathbf{D}^{\boldsymbol{\delta}}_{i^R}$ and the candidate restoration-action set $\mathbf{A}^{\boldsymbol{\delta}}_{i^R}$.
    \ForAll{$\rho\in\mathbf{A}^{\boldsymbol{\delta}}_{i^R}$}
        \State Generate the candidate next restored image-state with frozen \usir{$\mathcal{S}_{\theta}$}:
        \[
            \widehat{I}^{\boldsymbol{\delta}}_{i^R-1,\rho}
            =
            \usir{\mathcal{S}_{\theta}}\!\left(
            \widetilde{I}^{\boldsymbol{\delta}}_{i^R},\rho\right).
        \]
        \State Compute its image-quality label $Q(\rho)\!:=\!Q(\widehat{I}^{\boldsymbol{\delta}}_{i^R-1,\rho})$ using the IQA ensemble of Sec.~\ref{sec:aioiqa_ensemble}.
    \EndFor
    \State Predict per-action quality scores:
    \[
        \hat{\mathbf{q}}^{\boldsymbol{\delta}}_{i^R}
        =
        \aio{f_{\psi}}\!\left(
        \widetilde{I}^{\boldsymbol{\delta}}_{i^R},
        \mathcal{T}^{\boldsymbol{\delta}}\right)
        \in [0,1]^{|\mathbf{A}^{\boldsymbol{\delta}}_{i^R}|}.
    \]
    \State Update \aio{$f_{\psi}$} by the pairwise ranking loss matching predicted gaps to IQA gaps:
    \[
        \mathcal{L}_{\text{rank}}(\aio{f_{\psi}})
        =
        \frac{1}{|\mathcal{C}|}
        \sum_{(a,b)\in\mathcal{C}}
        \!\bigl(
        (\hat{\mathbf{q}}^{\boldsymbol{\delta}}_{i^R}[a]
        -\hat{\mathbf{q}}^{\boldsymbol{\delta}}_{i^R}[b])
        -(Q(\rho_a)-Q(\rho_b))
        \bigr)^2,
    \]
    where $\mathcal{C}=\{(a,b)\in\mathbf{A}^{\boldsymbol{\delta}}_{i^R}\times\mathbf{A}^{\boldsymbol{\delta}}_{i^R}:a\ne b\}$.
    \State Keep \usir{$\mathcal{S}_{\theta}$} frozen.
\EndFor
\State \Return \aio{$f_{\psi}$}.
\end{algorithmic}
\end{algorithm}

\subsection{Constructing $\mathcal{D}_{\text{ORTD}}^{\text{DiTTo}}$}

Algorithm~\ref{alg:construct_ditto_ortd} constructs the
DiTTo-based Optimal Restoration-action Trajectory Dataset
$\mathcal{D}_{\text{ORTD}}^{\text{DiTTo}}$ by combining the two
trained modules: \aio{$f_{\psi}$} selects the simulator-approximated
optimal candidate identifier
\[
\rho^{\boldsymbol{\delta}}_{i^R}
=
\operatorname*{argmax}_{\rho\in\mathbf{A}^{\boldsymbol{\delta}}_{i^R}}
\hat{\mathbf{q}}^{\boldsymbol{\delta}}_{i^R}[\rho]
\]
at each restoration index, and \usir{$\mathcal{S}_{\theta}$} applies
the corresponding restoration-action to produce the next restored
image-state.
For each training image-state, the procedure starts from the
observed multi-degraded image-state
$\widetilde{I}^{\boldsymbol{\delta}}_j:=I^{\boldsymbol{\delta}}_j$
with $\mathbf{D}^{\boldsymbol{\delta}}_j:=\mathbf{D}^{\boldsymbol{\delta}}$
and unrolls the restoration-action-trajectory from
$i^R{=}j$ down to $i^R{=}1$, adding the pair
$(\widetilde{I}^{\boldsymbol{\delta}}_{i^R},
A^R_{\rho^{\boldsymbol{\delta}}_{i^R}}(\cdot))$ to
$\mathcal{D}_{\text{ORTD}}^{\text{DiTTo}}$ at every step and
shrinking $\mathbf{D}^{\boldsymbol{\delta}}_{i^R}$ by one type after
each application.
The crucial property is that the entire trajectory is unrolled with
$\mathcal{O}(N^{\mathbf{D}})$ simulator steps per image, namely one
\aio{$f_{\psi}$} call and one \usir{$\mathcal{S}_{\theta}$} call per
restoration index, without any real restoration-expert evaluation,
which is what enables scalable supervision construction for the
DiTTo Agent.

\begin{algorithm}[h]
\caption{Constructing $\mathcal{D}_{\text{ORTD}}^{\text{DiTTo}}$ with \usir{$\mathcal{S}_{\theta}$} and \aio{$f_{\psi}$}}
\label{alg:construct_ditto_ortd}
\begin{algorithmic}[1]
\Require Trained \usir{$\mathcal{S}_{\theta}$}; trained \aio{$f_{\psi}$}; clean image-states $I_{\text{clean}}$; degradation universe $\mathbf{D}$; total instances $M$.
\Ensure $\mathcal{D}_{\text{ORTD}}^{\text{DiTTo}}$.

\State Initialize $\mathcal{D}_{\text{ORTD}}^{\text{DiTTo}}\leftarrow\varnothing$.
\For{$m=1,\ldots,M$}
    \State Sample a clean image-state $I_{\text{clean}}$.
    \State Sample a degradation-ordering $\boldsymbol{\delta}=(\delta_{i^D})_{i^D=1}^{j}$.
    \State Construct $\mathcal{T}^{\boldsymbol{\delta}}$ and obtain $I^{\boldsymbol{\delta}}_j$.
    \State Set $\widetilde{I}^{\boldsymbol{\delta}}_j:=I^{\boldsymbol{\delta}}_j$ and $\mathbf{D}^{\boldsymbol{\delta}}_j:=\mathbf{D}^{\boldsymbol{\delta}}$.
    \For{$i^R=j,j-1,\ldots,1$} \Comment{$|\mathbf{D}^{\boldsymbol{\delta}}_{i^R}|$ shrinks by one each iteration}
        \State Construct the candidate restoration-action set:
        \[
            \mathbf{A}^{\boldsymbol{\delta}}_{i^R}
            =
            \{(D,i^E_D):D\in\mathbf{D}^{\boldsymbol{\delta}}_{i^R},\,
            i^E_D\in\{1,\ldots,N^E_D\}\}.
        \]
        \State Predict per-action quality scores with \aio{$f_{\psi}$}:
        \[
            \hat{\mathbf{q}}^{\boldsymbol{\delta}}_{i^R}
            =
            \aio{f_{\psi}}\!\left(
            \widetilde{I}^{\boldsymbol{\delta}}_{i^R},
            \mathcal{T}^{\boldsymbol{\delta}}\right).
        \]
        \State Select the simulator-approximated optimal identifier:
        \[
            \rho^{\boldsymbol{\delta}}_{i^R}
            =
            \operatorname*{argmax}_{\rho\in\mathbf{A}^{\boldsymbol{\delta}}_{i^R}}
            \hat{\mathbf{q}}^{\boldsymbol{\delta}}_{i^R}[\rho].
        \]
        \State Add the ORTD pair:
        \[
            \mathcal{D}_{\text{ORTD}}^{\text{DiTTo}}
            \leftarrow
            \mathcal{D}_{\text{ORTD}}^{\text{DiTTo}}
            \cup
            \left\{
            \left(
            \widetilde{I}^{\boldsymbol{\delta}}_{i^R},
            A^R_{\rho^{\boldsymbol{\delta}}_{i^R}}(\cdot)
            \right)
            \right\}.
        \]
        \State Apply the selected restoration-action with \usir{$\mathcal{S}_{\theta}$}:
        \[
            \widetilde{I}^{\boldsymbol{\delta}}_{i^R-1}
            =
            \usir{\mathcal{S}_{\theta}}\!\left(
            \widetilde{I}^{\boldsymbol{\delta}}_{i^R},
            \rho^{\boldsymbol{\delta}}_{i^R}\right).
        \]
        \State Update the remaining-degradation type set by removing the type $D^{\boldsymbol{\delta}}_{i^R}$ specified by $\rho^{\boldsymbol{\delta}}_{i^R}$:
        \[
            \mathbf{D}^{\boldsymbol{\delta}}_{i^R-1}
            =
            \mathbf{D}^{\boldsymbol{\delta}}_{i^R}
            \setminus
            \{D^{\boldsymbol{\delta}}_{i^R}\}.
        \]
    \EndFor
\EndFor
\State \Return $\mathcal{D}_{\text{ORTD}}^{\text{DiTTo}}$.
\end{algorithmic}
\end{algorithm}

\section{$\cup$S-IR: Single-degradation Restoration Simulator Details}
\label{sec:usir_details}

This section provides the implementation details deferred from
Sec.~\ref{sec:method} of the main paper.
$\cup$S-IR is a single-degradation restoration simulator that
approximates the effect of an arbitrary restoration-action
without invoking real restoration-experts, enabling
$\mathcal{O}(N^{\mathbf{D}})$ construction of
$\mathcal{D}_{\text{ORTD}}^{\text{DiTTo}}$ via Algorithm~\ref{alg:construct_ditto_ortd}.

\subsection{Architecture}
\label{sec:usir_arch}

$\cup$S-IR is implemented as an action-conditioned latent diffusion
model with an SD3-style flow-matching backbone~\citep{esser2024scaling}.
Given the current restored image-state
$\widetilde{I}^{\boldsymbol{\delta}}_{i^R}$ and a candidate
restoration-action identifier $\rho=(D,i^E_D)\in\mathbf{A}^{\boldsymbol{\delta}}_{i^R}$,
$\cup$S-IR predicts the next restored image-state
$\widehat{I}^{\boldsymbol{\delta}}_{i^R-1,\rho}=
\mathcal{S}_{\theta}(\widetilde{I}^{\boldsymbol{\delta}}_{i^R},\rho)$.
We use a pretrained SD3.5-medium VAE
($H{=}W{=}512$ to $32{\times}32$ latent, $C_z{=}16$) frozen
throughout, and train a $12$-layer MM-DiT over the latent space
($\text{patch}{=}2$, heads${=}10$, head dim${=}64$,
$\text{pos\_embed\_max}{=}96$).
The candidate restoration-action identifier $\rho$ is embedded
together with the indicator
$\mathbf{1}[\mathbf{D}^{\boldsymbol{\delta}}_{i^R}]$ over remaining
degradation types and the order-rank vector
$\mathbf{o}^{\boldsymbol{\delta}}_{i^R}$ summarising
$\mathcal{T}^{\boldsymbol{\delta}}$, and injected into every DiT
block via cross-attention.
The VAE remains frozen.

\subsection{Action-Conditioned Clean/Degraded Feature Mixing}
\label{sec:usir_mix}

A naive single-degradation simulator that simply maps degraded
to clean loses the non-target degradations that must be preserved.
We therefore form two parallel feature streams:
a clean-conditioned stream $\mathbf{h}_{\text{clean}}$ that provides
restoration cues for the target degradation type $D$, and a
degraded-conditioned stream $\mathbf{h}_{\text{deg}}$ that preserves
the remaining degradations
$\mathbf{D}^{\boldsymbol{\delta}}_{i^R}\setminus\{D\}$.
The simulator output is an action-conditioned mixture of the two,
gated per frequency band as detailed below.

\subsection{Adaptive Frequency-Band Mixing}
\label{sec:usir_freq}

We split each feature into $K$ frequency bands using a 2D-DCT-based
bandpass and assign band-wise gates
$g^{(k)}_{\rho}\in[0,1]$ predicted from $\rho$:
\[
\mathbf{h}_{\text{mix}}^{(k)}=
g^{(k)}_{\rho}\,\mathbf{h}_{\text{clean}}^{(k)}+
(1-g^{(k)}_{\rho})\,\mathbf{h}_{\text{deg}}^{(k)},
\qquad k=1,\ldots,K,
\]
with $g^{(k)}_{\rho}=\sigma(\mathrm{MLP}_{\rho}^{(k)}([\mathbf{e}_\rho,\mathbf{e}_{D}]))$.
The intuition is that different degradation types occupy different
frequency bands (e.g., low-light dominantly affects illumination/low
band, sensor noise the high band), so an action-conditioned
band-wise gate can route restoration cues to the correct band while
leaving the rest of the spectrum intact.
We use $K{=}4$ throughout the main paper; the ablation in
Sec.~\ref{sec:usir_K_ablation} sweeps $K$.

\subsection{Training Objective}
\label{sec:usir_loss}

We adopt the SD3-style logit-normal sigma sampling and flow-matching
weighting~\citep{esser2024scaling}.
Let $\mathbf{z}_t=(1-\sigma)\,\mathbf{z}_{\text{target}}+\sigma\,\boldsymbol{\epsilon}$
with $\boldsymbol{\epsilon}\!\sim\!\mathcal{N}(0,\mathbf{I})$, and
$\widehat{\mathbf{z}}_0=\mathbf{z}_t-\sigma\,\mathbf{v}_\theta$ the
$\mathbf{z}_0$-prediction reconstructed from the predicted velocity.
The base loss is a sigma-weighted Charbonnier on
$\widehat{\mathbf{z}}_0$:
\[
\mathcal{L}_{\text{base}}=
\mathbb{E}\!\left[
w(\sigma)\,
\sqrt{(\widehat{\mathbf{z}}_0-\mathbf{z}_{\text{target}})^2+\epsilon^2}
\right],
\quad \epsilon{=}10^{-3}.
\]
The action-conditioned mixing in Sec.~\ref{sec:usir_mix} can collapse
to either stream when one degradation dominates the loss.
To mitigate this, we additionally apply a small-active sparse mask:
let $\boldsymbol{\Delta}=|\mathbf{z}_{\text{target}}-\mathbf{z}_{\text{deg}}|$
be the per-pixel target-to-degraded delta in the latent.
We mark a position as ``small-delta but active'' if it lies in the
lower-$q$ quantile of $\boldsymbol{\Delta}$ \emph{and} the upper-$q_a$
quantile of $|\mathbf{z}_{\text{target}}|$
($q{=}0.3$, $q_a{=}0.1$).
This gives a binary mask $\mathbf{m}$ that highlights regions which
are visually content-rich but should not change under the chosen
restoration-action (i.e., regions of non-target degradations that must
be preserved).
The sparse Charbonnier loss is
$\mathcal{L}_{\text{sparse}}=
\frac{1}{\|\mathbf{m}\|_1}\sum w(\sigma)\,\mathbf{m}\odot
\sqrt{(\widehat{\mathbf{z}}_0-\mathbf{z}_{\text{target}})^2+\epsilon^2}$.
The full objective is
$\mathcal{L}=\mathcal{L}_{\text{base}}+\lambda_s\,\mathcal{L}_{\text{sparse}}$
with $\lambda_s{=}1.0$.

\subsection{Implementation}
\label{sec:usir_impl}

We train $\cup$S-IR on the clean image-states described in
Sec.~\ref{sec:data}, with on-the-fly degradation synthesis
(Sec.~\ref{sec:data_degsynth}).
At each step we sample $j\!\in\!\{2,\ldots,6\}$, build
$\mathcal{T}^{\boldsymbol{\delta}}$ by sequentially composing the
degradation-actions $A^D_{D_{\delta_{i^D}}}(\cdot)$, then sample one
$i^R\in\{1,\ldots,j\}$ and one candidate
$\rho\in\mathbf{A}^{\boldsymbol{\delta}}_{i^R}$ to form the
prediction target
$\widetilde{I}^{\boldsymbol{\delta}}_{i^R-1}$.
We use AdamW ($\beta_1{=}0.9,\beta_2{=}0.95$, weight decay $10^{-2}$),
peak LR $5\!\times\!10^{-4}$ with cosine decay to $5\!\times\!10^{-6}$,
warmup ratio $1/5$, and gradient clipping at $1.0$.
Mixed precision (fp16) with FlowMatchEulerDiscrete scheduling is used
throughout.
Training runs on $2{\times}$B200.

\subsection{Ablation: Number of Frequency Bands $K$}
\label{sec:usir_K_ablation}

The choice of $K$ controls how finely $\cup$S-IR can route
restoration cues across the spectrum.
We sweep $K\!\in\!\{1,2,4,8\}$ keeping all other settings fixed and
evaluate the single-degradation prediction
$\widetilde{I}^{\boldsymbol{\delta}}_{i^R}\!\to\!\widetilde{I}^{\boldsymbol{\delta}}_{i^R-1}$
on a held-out set, reporting PSNR / LPIPS / MANIQA / MUSIQ averaged
over $j\!\in\!\{2,\ldots,6\}$ and all candidates
$\rho\in\mathbf{A}^{\boldsymbol{\delta}}_{i^R}$.

Tab.~\ref{tab:usir_K_ablation} reveals a clear non-monotonic trend across all four metrics, confirming that frequency-band granularity must be matched to the spectral structure of the degradation universe.
At $K{=}1$, the band-wise gate degenerates into a single global gate, which forces the simulator to apply the same clean/degraded mixing ratio to every frequency component.
This is fundamentally misaligned with the degradation universe $\mathbf{D}$, since low-frequency-dominant degradations (fog, low-light) and high-frequency-dominant degradations (sensor noise, defocus blur) require opposite mixing decisions.
The result is a measurable drop in both PSNR and LPIPS, indicating that the simulator either under-restores the target degradation or over-modifies non-target frequency bands.

Increasing $K$ from $1$ to $4$ progressively allocates separate gates to each spectral region, allowing the simulator to leave non-target bands untouched while restoring the target.
The improvement is most visible on no-reference metrics (MANIQA, MUSIQ), which are particularly sensitive to spurious frequency-domain artifacts introduced by mismatched gates.
At $K{=}4$, the four bands align well with the dominant spectral signatures of the six degradation types in $\mathbf{D}$, and the simulator achieves the best trade-off across all metrics.

Pushing further to $K{=}8$ over-fragments the spectrum: each band carries less restoration cue, and adjacent bands begin to compete for the same gate signal predicted from $\rho$.
This is especially harmful for low-frequency restoration cues (fog removal, illumination correction), where the relevant signal spans a wide low-band region that $K{=}8$ splits into multiple sub-bands.
PSNR drops accordingly, and LPIPS regresses as the simulator introduces band-boundary artifacts.
$K{=}4$ therefore represents the operating point where spectral granularity matches degradation diversity in $\mathbf{D}$ without over-fragmenting the restoration signal.

\begin{table}[h]
\centering
\caption{Ablation on the number of frequency bands $K$ in $\cup$S-IR.
We report single-degradation prediction quality
($\widetilde{I}^{\boldsymbol{\delta}}_{i^R}\!\to\!\widetilde{I}^{\boldsymbol{\delta}}_{i^R-1}$)
averaged over $j\!\in\!\{2,\ldots,6\}$ and all candidate identifiers
$\rho\in\mathbf{A}^{\boldsymbol{\delta}}_{i^R}$.}
\label{tab:usir_K_ablation}
\small
\begin{tabular}{lcccc}
\toprule
$K$ & PSNR\,$\uparrow$ & LPIPS\,$\downarrow$ & MANIQA\,$\uparrow$ & MUSIQ\,$\uparrow$ \\
\midrule
1 & 21.69 & 0.2798 & 0.2109 & 38.48 \\
2 & 22.31 & 0.2198 & 0.2160 & 39.20 \\
\textbf{4} & 24.25 & 0.1891 & 0.2287 & 40.44 \\
8 & 23.82 & 0.2034 & 0.2274 & 39.96 \\
\bottomrule
\end{tabular}
\end{table}

\section{AiO-IQA: All-in-One Restoration-Action Scoring Details}
\label{sec:aioiqa_details}

\subsection{Architecture}
\label{sec:aioiqa_arch}

AiO-IQA $f_{\psi}$ takes the current restored image-state
$\widetilde{I}^{\boldsymbol{\delta}}_{i^R}$ and the
degradation-action-trajectory $\mathcal{T}^{\boldsymbol{\delta}}$
as input, and outputs a candidate-aligned score vector
$\hat{\mathbf{q}}^{\boldsymbol{\delta}}_{i^R}\!\in\![0,1]^{|\mathbf{A}^{\boldsymbol{\delta}}_{i^R}|}$
under a fixed enumeration of
$\mathbf{A}^{\boldsymbol{\delta}}_{i^R}$
(Sec.~\ref{sec:notation}).
Concretely, we encode the latent
$\mathbf{z}=\mathrm{VAE}(\widetilde{I}^{\boldsymbol{\delta}}_{i^R})$
with a multi-scale convolutional encoder (three scales pooled to
$\{4{\times}4,2{\times}2,1{\times}1\}$, each projected to $256$-dim
and concatenated to a $768$-dim feature),
condition it on a FiLM~\citep{perez2018film} produced from the concatenation
of the indicator $\mathbf{1}[\mathbf{D}^{\boldsymbol{\delta}}_{i^R}]$
and the order-rank vector $\mathbf{o}^{\boldsymbol{\delta}}$
(both summarising $\mathcal{T}^{\boldsymbol{\delta}}$), and decode
to the candidate-aligned score with an MLP head.
Entries corresponding to types not in
$\mathbf{D}^{\boldsymbol{\delta}}_{i^R}$ are masked to $-\infty$ at
inference, which makes the
$\operatorname*{argmax}_{\rho\in\mathbf{A}^{\boldsymbol{\delta}}_{i^R}}$
in Algorithm~\ref{alg:construct_ditto_ortd} (line 10) safe.

\subsection{IQA Metric Ensemble}
\label{sec:aioiqa_ensemble}

The supervisory $Q(\cdot)$ in Algorithm~\ref{alg:train_aio_iqa}
(line 12) is a fixed scalar combination of full-reference and
no-reference IQA metrics computed on the
$\cup$S-IR-generated next restored image-state
$\mathcal{S}_{\theta}(\widetilde{I}^{\boldsymbol{\delta}}_{i^R},\rho)$:
\[
Q=
w_{\text{NR}}\bar{Q}_{\text{NR}}
+w_{\text{PSNR}}Q_{\text{PSNR}}
+w_{\text{SSIM}}Q_{\text{SSIM}},
\quad w_{\text{NR}}{=}0.5,\ w_{\text{PSNR}}{=}w_{\text{SSIM}}{=}0.25,
\]
where $\bar{Q}_{\text{NR}}$ is the mean of MANIQA, MUSIQ, CLIP-IQA,
and a sigmoid-transformed NIQE
$e^{-\text{NIQE}/10}$, all normalised to $[0,1]$.
$Q_{\text{PSNR}}$ and $Q_{\text{SSIM}}$ are computed against the
synthetic reference
$\widetilde{I}^{\boldsymbol{\delta},*}_{i^R-1}$ which is known by
construction during synthesis (see Sec.~\ref{sec:data}).

\subsection{Score Normalization and Ranking Objective}
\label{sec:aioiqa_loss}

Because absolute IQA values vary across instances and restoration
indices, we supervise $f_{\psi}$ with a per-instance \emph{pairwise}
ranking objective rather than a regression to absolute scores.
Let $Q(\rho_a),Q(\rho_b)$ be the IQA scores of two candidates at the
same $(\widetilde{I}^{\boldsymbol{\delta}}_{i^R},\mathcal{T}^{\boldsymbol{\delta}})$.
We minimise
\[
\mathcal{L}_{\text{rank}}=
\frac{1}{|\mathcal{C}|}
\sum_{(a,b)\in\mathcal{C}}
\bigl(
(\hat{\mathbf{q}}^{\boldsymbol{\delta}}_{i^R}[a]
-\hat{\mathbf{q}}^{\boldsymbol{\delta}}_{i^R}[b])
-(Q(\rho_a)-Q(\rho_b))
\bigr)^2,
\]
where $\mathcal{C}$ enumerates all candidate pairs
$(a,b)\in\mathbf{A}^{\boldsymbol{\delta}}_{i^R}\times\mathbf{A}^{\boldsymbol{\delta}}_{i^R}$
with $a\ne b$.
This pairwise regression loss matches both the rank ordering and the
relative gaps between candidates, which is exactly what is needed by
the $\operatorname*{argmax}$ selection step in
Algorithm~\ref{alg:construct_ditto_ortd}.

\subsection{Implementation}

We train $f_{\psi}$ with $\mathcal{S}_{\theta}$ frozen
(Algorithm~\ref{alg:train_aio_iqa}), AdamW with peak LR
$5\!\times\!10^{-4}$, cosine schedule on $2{\times}$B200, mixed precision fp16.
We additionally apply a curriculum on $j$ ($2$ to $6$) and
per-degradation severity to stabilise early training.

\subsection{Validation: Per-Action Ranking Accuracy}
\label{sec:aioiqa_val}

We measure how well $f_{\psi}$ recovers the $Q$-induced ordering on
a held-out validation set, reporting Recall@1, the fraction of
instances where the AiO-IQA-selected
$\rho^{\boldsymbol{\delta}}_{i^R}=
\operatorname*{argmax}_{\rho}\hat{\mathbf{q}}^{\boldsymbol{\delta}}_{i^R}[\rho]$
matches $\operatorname*{argmax}_{\rho}Q(\rho)$, and the Spearman
rank correlation $\rho_s$ over $\mathbf{A}^{\boldsymbol{\delta}}_{i^R}$.
Tab.~\ref{tab:aioiqa_val} reports both metrics, broken down by
$j=|\mathbf{D}^{\boldsymbol{\delta}}|\in\{2,\ldots,6\}$ to expose
the difficulty of larger candidate sets.

The two metrics together characterize complementary aspects of $f_{\psi}$: Recall@1 measures whether AiO-IQA picks the same top candidate as the IQA ensemble (the only choice that matters for Algorithm~\ref{alg:construct_ditto_ortd}, which uses $\operatorname*{argmax}$ selection), while Spearman captures how faithfully $f_{\psi}$ reproduces the full ranking over $\mathbf{A}^{\boldsymbol{\delta}}_{i^R}$.
Recall@1 stays above $0.80$ for $j\!\in\!\{2,3,4,5\}$, which directly validates the $\mathcal{O}(N^{\mathbf{D}})$ ORTD construction in Algorithm~\ref{alg:construct_ditto_ortd}: in over four out of five instances, $f_{\psi}$ selects the same simulator-approximated optimal identifier $\rho^{\boldsymbol{\delta}}_{i^R}$ that an exhaustive IQA-ensemble evaluation would have picked, but at a fraction of the cost.

The drop at $j{=}6$ (Recall@1 from $0.803$ to $0.758$) reflects the combinatorial growth of the candidate set $\mathbf{A}^{\boldsymbol{\delta}}_{i^R}$: as more degradation types are present, more (degradation type, restoration-expert) pairs become viable at each restoration index, and small differences between top candidates become harder to discriminate.
Importantly, Spearman remains essentially flat ($0.747$ at $j{=}6$ vs.\ $0.767$ at $j{=}2$), indicating that $f_{\psi}$ continues to reproduce the global IQA ordering even when the top-1 decision becomes harder.
This is the desirable failure mode for ORTD construction: when the top-1 selection misses the IQA-best candidate, $f_{\psi}$ tends to pick a near-best alternative rather than a low-quality one, so the resulting trajectory in $\mathcal{D}_{\text{ORTD}}^{\text{DiTTo}}$ remains close to the optimal trajectory in IQA terms.
The remaining gap at $j{=}6$ is one of the reasons for the Stage~2 ORA refinement, which corrects residual simulator-to-expert deviations on a small expert-executed subset.

\begin{table}[h]
\centering
\caption{AiO-IQA per-action ranking accuracy on a held-out
validation set. We report Recall@1 (top-1 match with the
IQA-best candidate) and Spearman over
$\mathbf{A}^{\boldsymbol{\delta}}_{i^R}$,
broken down by the number of involved degradations
$j=|\mathbf{D}^{\boldsymbol{\delta}}|$.}
\label{tab:aioiqa_val}
\small
\begin{tabular}{lcccccc}
\toprule
$j$ & 2 & 3 & 4 & 5 & 6 & avg.\\
\midrule
Recall@1\,$\uparrow$ & 0.823 & 0.810 & 0.805 & 0.803 & 0.758 & 0.736 \\
Spearman\,$\uparrow$ & 0.767 & 0.751 & 0.750 & 0.752 & 0.747 & 0.749 \\
\bottomrule
\end{tabular}
\end{table}

\section{DiTTo Agent Training Details}
\label{sec:agent_details}

\subsection{VLM Backbone and LoRA Configuration}

We instantiate $\mathit{Agent}_{DP+OR}$ on top of
Qwen3-VL-8B-Thinking and fine-tune with LoRA on the attention and
MLP projection modules ($r{=}16$, $\alpha{=}32$, dropout $0.05$).
To preserve \verb|<think>| blocks during multi-turn fine-tuning
we bypass the default \verb|apply_chat_template| and construct
training sequences manually so that reasoning traces are part of
the supervised target.

\subsection{Multi-Turn Tool-Use Conversation Format}

Each instance in $\mathcal{D}_{\text{ORTD}}^{\text{DiTTo}}$
is converted into a multi-turn conversation in which, at each
restoration index $i^R$, the assistant
(i) reasons about the remaining-degradation type set
$\mathbf{D}^{\boldsymbol{\delta}}_{i^R}$ inside a \verb|<think>|
block, then
(ii) emits a JSON-based tool call specifying the chosen
restoration-action
$\rho^{\boldsymbol{\delta}}_{i^R}=
(D^{\boldsymbol{\delta}}_{i^R},i^E_{D^{\boldsymbol{\delta}}_{i^R}})$:
\begin{verbatim}
{"action": "<degradation_type>", "model": "<expert_id>"}
\end{verbatim}
The tool result (i.e., the next restored image-state
$\widetilde{I}^{\boldsymbol{\delta}}_{i^R-1}$) is fed back as the
next user turn, and the loop repeats until $i^R{=}0$.

\subsection{Stage 1 SFT}

We train Stage 1 SFT on
$\mathcal{D}_{\text{ORTD}}^{\text{DiTTo}}$ with AdamW
(peak LR $1\!\times\!10^{-4}$, cosine to $1\!\times\!10^{-5}$,
warmup ratio $0.05$), batch size $1$ per device with gradient
accumulation $8$, and max sequence length $8192$.
We use greedy decoding at inference for structured-JSON parse
stability.

\subsection{Stage 2 ORA (Order-aware Restoration Alignment)}

\paragraph{Objective.}
ORA is a DPO-style objective applied to the decomposed planning
axes (DP, OR, Tool) introduced in the main paper.
Let $\pi_\theta$ and $\pi_{\text{ref}}$ be the policy and reference
models, and let $(y^c,y^r)$ be a chosen/rejected response pair
sharing the same prompt $x$.
We define three disjoint token masks
$\mathbf{m}^{(\text{DP})}, \mathbf{m}^{(\text{OR})},
\mathbf{m}^{(\text{Tool})}$ that select tokens belonging to the
Degradation Perception-Reasoning, Order-aware Restoration, and
JSON-based tool-call axes of the response, respectively.
The per-axis log-ratio is
\[
r^{(a)}_\theta(y\mid x)=
\sum_t m^{(a)}_t\bigl[
\log\pi_\theta(y_t\mid x,y_{<t})-\log\pi_{\text{ref}}(y_t\mid x,y_{<t})
\bigr],\quad a\in\{\text{DP},\text{OR},\text{Tool}\},
\]
and the ORA loss is a weighted sum of axis-wise DPO terms
\[
\mathcal{L}_{\text{ORA}}=
-\sum_{a\in\{\text{DP},\text{OR},\text{Tool}\}}\!\!\!\!
\lambda_a\,
\log\sigma\!\left(\beta\bigl[r^{(a)}_\theta(y^c\mid x)
-r^{(a)}_\theta(y^r\mid x)\bigr]\right),
\]
with $\beta{=}0.1$ and
$(\lambda_{\text{DP}},\lambda_{\text{OR}},\lambda_{\text{Tool}})
=(1.0,1.0,0.5)$.
This decomposition addresses the dilution problem of generic
response-level DPO: chosen and rejected trajectories share most
reasoning and JSON-template tokens, so a single response-level
margin assigns most of the gradient to tokens that are not
informative about $\mathbf{D}^{\boldsymbol{\delta}}$ or
$\boldsymbol{\rho}^{\boldsymbol{\delta}}$.

\paragraph{Chosen / rejected construction.}
\textbf{Chosen} restoration-action-trajectories come from the small expert-executed
subset $\mathcal{D}_{\text{ORTD}}^{\text{Expert}}$ produced by a
greedy expert search described in Sec.~\ref{sec:data_dpo}, where at
each $i^R$ both the degradation type and the restoration-expert
index are picked to maximise a combined IQA score using the real
restoration-experts.
\textbf{Rejected} restoration-action-trajectories are generated by the DiTTo Simulator
on the \emph{same} input $\widetilde{I}^{\boldsymbol{\delta}}_j$
using AiO-IQA-driven greedy ordering on $\mathcal{S}_{\theta}$
rather than real restoration-experts, plus two augmentations:
(i) a \emph{failure-injection} variant that randomly perturbs the
greedy choice with a small probability, and
(ii) a \emph{format-violation} variant that breaks the JSON-based
tool-call schema.
We crucially use the simulator-rendered restored image-state as
the rejected context image even when the chosen trajectory uses an
expert-rendered restored image-state, which prevents the chosen
pair from trivially winning on log-prob due to higher visual quality
of its conditioning image.
Pairs are weighted equally; the format-violation pair contributes
only to the Tool axis.

\paragraph{Hyperparameters.}
ORA is run on the small expert-executed subset
$\mathcal{D}_{\text{ORTD}}^{\text{Expert}}$
($j\!\in\!\{2,3,4,5\}$) with
AdamW (LR $5\!\times\!10^{-6}$, cosine schedule,
batch $1\!\times\!\text{accum}\,8$), reference-model frozen at the
$\mathcal{W}^{\text{SFT}}_{\text{DiTTo}}$ initialization.
We monitor per-axis margins and select the
final checkpoint as $\mathcal{W}^{\text{ORA}}_{\text{DiTTo}}$
based on validation
restoration quality on a held-out
$j\!\in\!\{2,\ldots,5\}$ subset.

\section{Restoration-Expert Pool}
\label{sec:expert_pool}

DiTTo Agent uses the same restoration-expert pool as
JarvisIR~\citep{lin2025jarvisir} (the configuration reported in
Tab.~\ref{perception_table} of the main paper).
$\star$DiTTo Agent extends this pool with recent state-of-the-art
restoration-experts (\verb|ipcdehaze| for fog and \verb|csud| for
rain streaks, both CVPR 2025) to demonstrate plug-and-play scalable
extensibility.
The full pool with per-type cardinalities $N^E_{D_n}$ is listed in
Tab.~\ref{tab:expert_pool}.
All restoration-experts are loaded from official public weights and
run at $512{\times}512$ unless their inference pipeline mandates
otherwise.

\begin{table}[h]
\centering
\caption{Restoration-expert pool. ``DiTTo Agent'' uses the same
pool as JarvisIR; ``$\star$DiTTo Agent'' is the extended pool used
in Tab.~\ref{perception_table} of the main paper.
$N^E_{D_n}$ denotes the pool size of degradation type $D_n$.}
\label{tab:expert_pool}
\small
\begin{tabular}{llcc}
\toprule
$D_n$ & Restoration-experts & $N^E_{D_n}$ (DiTTo) & $N^E_{D_n}$ ($\star$DiTTo) \\
\midrule
fog (haze)        & ridcp, kanet, \textit{ipcdehaze}                & 2 & 3 \\
rain streaks        & idt, turbo\_rain, s2former, \textit{csud}       & 3 & 4 \\
snow                & turbo\_snow, snowmaster                       & 2 & 2 \\
low-light enhance.  & retinexformer\_fivek, hvicidnet, lightdiff    & 3 & 3 \\
sensor noise        & scunet                                        & 1 & 1 \\
super-resolution    & real\_esrgan, \textit{AdcSR}                                        & 1 & 2 \\
defocus blur        & \textit{drbnet}                                        & 0 & 1 \\
\bottomrule
\end{tabular}
\end{table}

\section{Training Data Construction}
\label{sec:data}

\subsection{Clean Image-State Sources}

We assemble training clean image-states $I_{\text{clean}}$ from
DIV2K-train~\cite{agustsson2017ntire}, DIV8K~\cite{gu2019div8k},
Flickr8K~\cite{hodosh2013framing}, Flickr2K~\cite{lim2017enhanced},
and NKUSR8K~\cite{duan2025dit4sr}.
All images are randomly cropped to $512{\times}512$ (with bilinear
up-sampling for sub-$512$ images) before degradation synthesis.

\subsection{Degradation Synthesis Pipeline}
\label{sec:data_degsynth}

For each clean image-state $I_{\text{clean}}$ we sample a
tuple length $j\!\sim\!\mathcal{U}\{2,\ldots,6\}$, sample
$\boldsymbol{\delta}$ as a random ordering of $j$ distinct types
from $\mathbf{D}$ (subject to the exclusion rule below), and
construct the degradation-action-trajectory
$\mathcal{T}^{\boldsymbol{\delta}}$ by sequentially composing the
degradation-actions $A^D_{D_{\delta_{i^D}}}(\cdot)$.
Per-degradation severity is sampled as
$\ell\!\sim\!\mathcal{U}(0.1,\ell_{\max}(j))$ with
$\ell_{\max}(j)$ decreasing from $0.5$ at $j{=}2$ to $0.2$ at
$j{=}6$ to keep visible content non-trivial under heavy compositions.
Tab.~\ref{tab:deg_params} summarises the parameter ranges.
We additionally exclude two visually conflicting compositions:
$\{\text{snow},\text{rain streaks}\}$ and
$\{\text{snow},\text{low-light}\}$.

\begin{table}[h]
\centering
\caption{Degradation synthesis parameter ranges.
$\ell\in[0.1,\ell_{\max}(j)]$ is the per-degradation severity
parameter passed to $A^D_{D_n}(\cdot)$.}
\label{tab:deg_params}
\small
\begin{tabular}{ll}
\toprule
$D_n$ & Internal parameter range driven by $\ell$ \\
\midrule
fog                  & atmospheric light $A\!\in\![0.7,0.95]$, transmission $t\!\in\![0.3,0.9]$ \\
rain streaks         & streak density / length / angle scaled by $\ell$ \\
snow                 & snow particle density / size scaled by $\ell$ \\
low-light enhance.   & gamma $\gamma\!\in\![1.5,4.0]$ + multiplicative gain $\!\in\![0.05,0.5]$ \\
sensor noise         & Gaussian $\sigma\!\in\![5,50]/255$ + Poisson shot noise \\
defocus blur         & disk-kernel radius $\!\in\![1,7]$ \\
\bottomrule
\end{tabular}
\end{table}

\subsection{$\mathcal{D}_{\text{ORTD}}^{\text{DiTTo}}$ Statistics}
\label{sec:data_ortd}

$\mathcal{D}_{\text{ORTD}}^{\text{DiTTo}}$ contains
trajectories balanced across $j\!\in\!\{2,3,4,5,6\}$, each yielding $j$ ORTD pairs
$(\widetilde{I}^{\boldsymbol{\delta},*}_{i^R},
A^R_{\rho^{\boldsymbol{\delta},*}_{i^R}}(\cdot))$.
Trajectories are produced by Algorithm~\ref{alg:construct_ditto_ortd}
with the trained $\mathcal{S}_{\theta}$ and $f_{\psi}$.

\subsection{$\mathcal{D}_{\text{ORTD}}^{\text{Expert}}$ Subset for ORA}
\label{sec:data_dpo}

The ORA subset of $\mathcal{D}_{\text{ORTD}}^{\text{Expert}}$
is balanced across $j\!\in\!\{2,3,4,5\}$.
For each instance, the chosen restoration-action-trajectory is
produced by a greedy search over candidate identifiers
$\rho=(D,i^E_D)\in\mathbf{A}^{\boldsymbol{\delta}}_{i^R}$ at every
$i^R$ on the real restoration-expert pool, scored by the combined
IQA in Sec.~\ref{sec:aioiqa_ensemble}; the rejected
restoration-action-trajectory is produced by the DiTTo Simulator
on the same input with AiO-IQA-driven ordering, plus the
failure-injection and format-violation augmentations described in
Sec.~\ref{sec:agent_details}.

\subsection{Dataset Release}
\label{sec:data_release}

For transparency and reproducibility, we provide a representative inspection subset of $\mathcal{D}*{\text{ORTD}}^{\text{DiTTo}}$ (approximately 100 trajectories balanced across $j!\in!{2,3,4,5,6}$), together with the corresponding multi-turn tool-use conversations described in Sec.~\ref{sec:agent_details}.
This subset facilitates understanding and verification of the data format, ORTD pair structure, and SFT conversation template.
The complete $\mathcal{D}*{\text{ORTD}}^{\text{DiTTo}}$ dataset, along with the trained $\mathcal{S}*{\theta}$ and $f*{\psi}$ checkpoints required to extend the dataset to newly introduced restoration-experts via Algorithm~\ref{alg:construct_ditto_ortd}, will be made publicly available in a future release.


\subsection{ORTD Example}
\label{sec:data_example}

\begin{figure}[h]
\centering
\includegraphics[width=\linewidth]{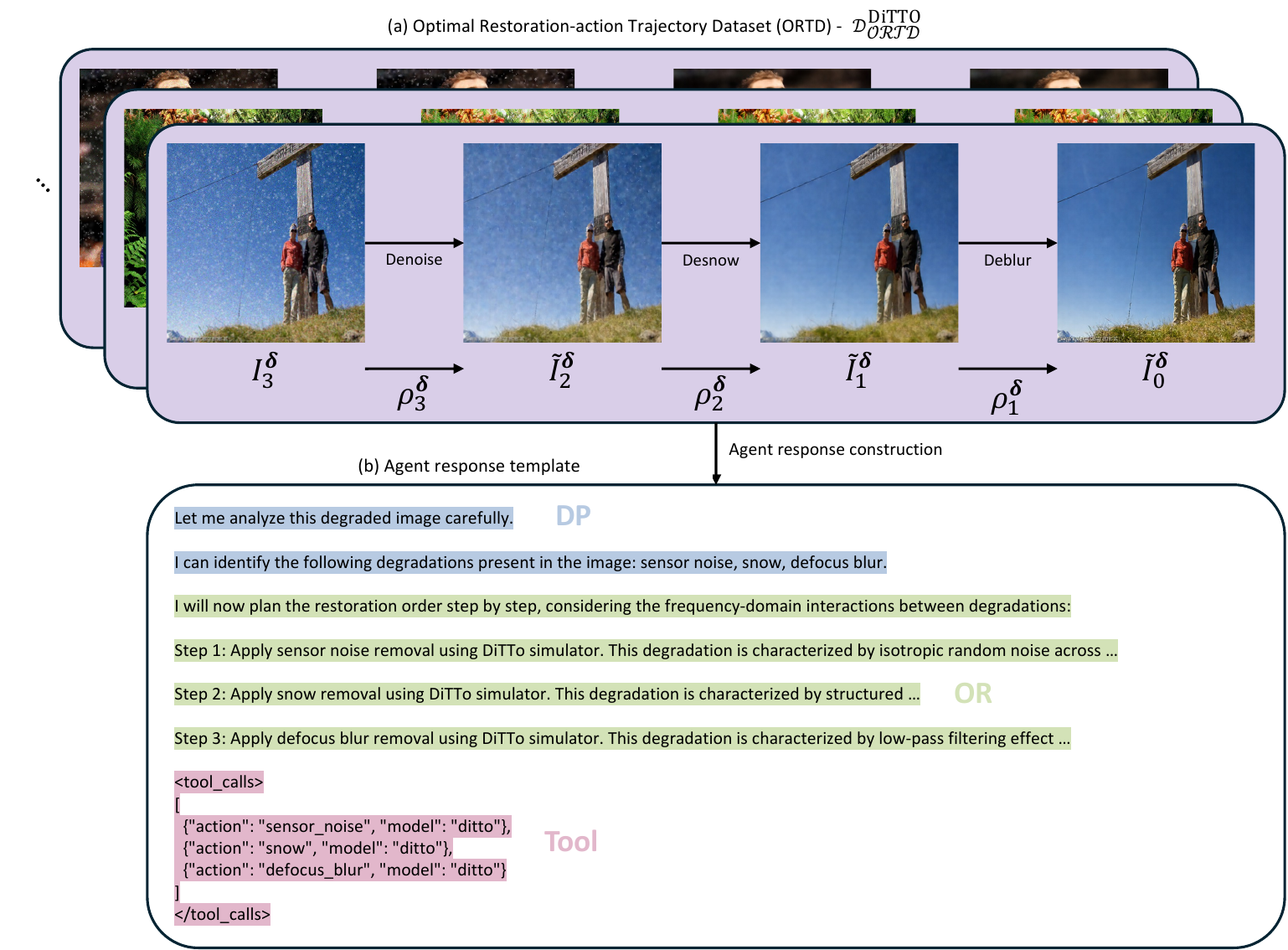}
\caption{An ORTD example with $j{=}3$. 
(a) The simulator-generated optimal restoration-action-trajectory 
$(\widetilde{I}^{\boldsymbol{\delta},*}_{i^R})_{i^R=3}^{0}$ in 
$\mathcal{D}_{\text{ORTD}}^{\text{DiTTo}}$. 
(b) The corresponding agent response, decomposed into 
DP (Degradation Perception-Reasoning), OR (Order-aware Restoration), 
and Tool (JSON-based tool call) axes used in ORA.}
\label{fig:ortd_example}
\end{figure}

Fig.~\ref{fig:ortd_example} shows one sampled instance with
$j{=}3$, including the synthesised observed multi-degraded
image-state $I^{\boldsymbol{\delta}}_3$, the simulator-generated
optimal restoration-action-trajectory
$(\widetilde{I}^{\boldsymbol{\delta},*}_{i^R})_{i^R=3}^{0}$
under
$\boldsymbol{\rho}^{\boldsymbol{\delta},*}=
((D_{n_1},\cdot),(D_{n_2},\cdot),(D_{n_3},\cdot))$,
and the corresponding multi-turn tool-use conversation that becomes
the SFT target.

\clearpage
\section{Qualitative Comparison}
\label{sec:qualitative}
\begin{figure}[h]
    \centering
    \includegraphics[width=0.85\linewidth]{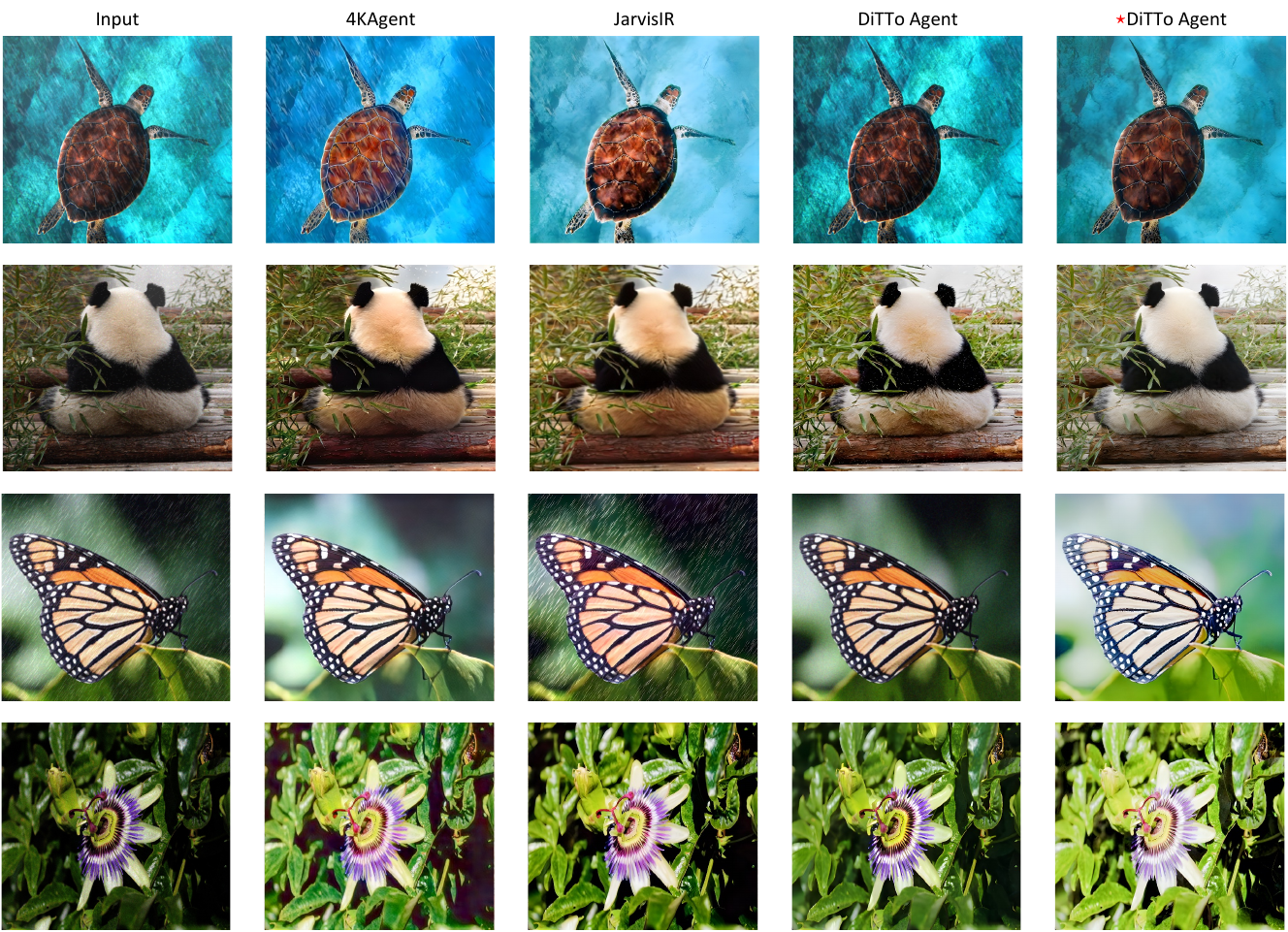}
    \caption{
    Additional qualitative comparisons on multi-degraded inputs with $j\!\in\!\{2,3\}$ concurrent degradations.}
    \label{fig:qualitative_low_j}
\end{figure}
\begin{figure}[h]
    \centering
    \includegraphics[width=0.85\linewidth]{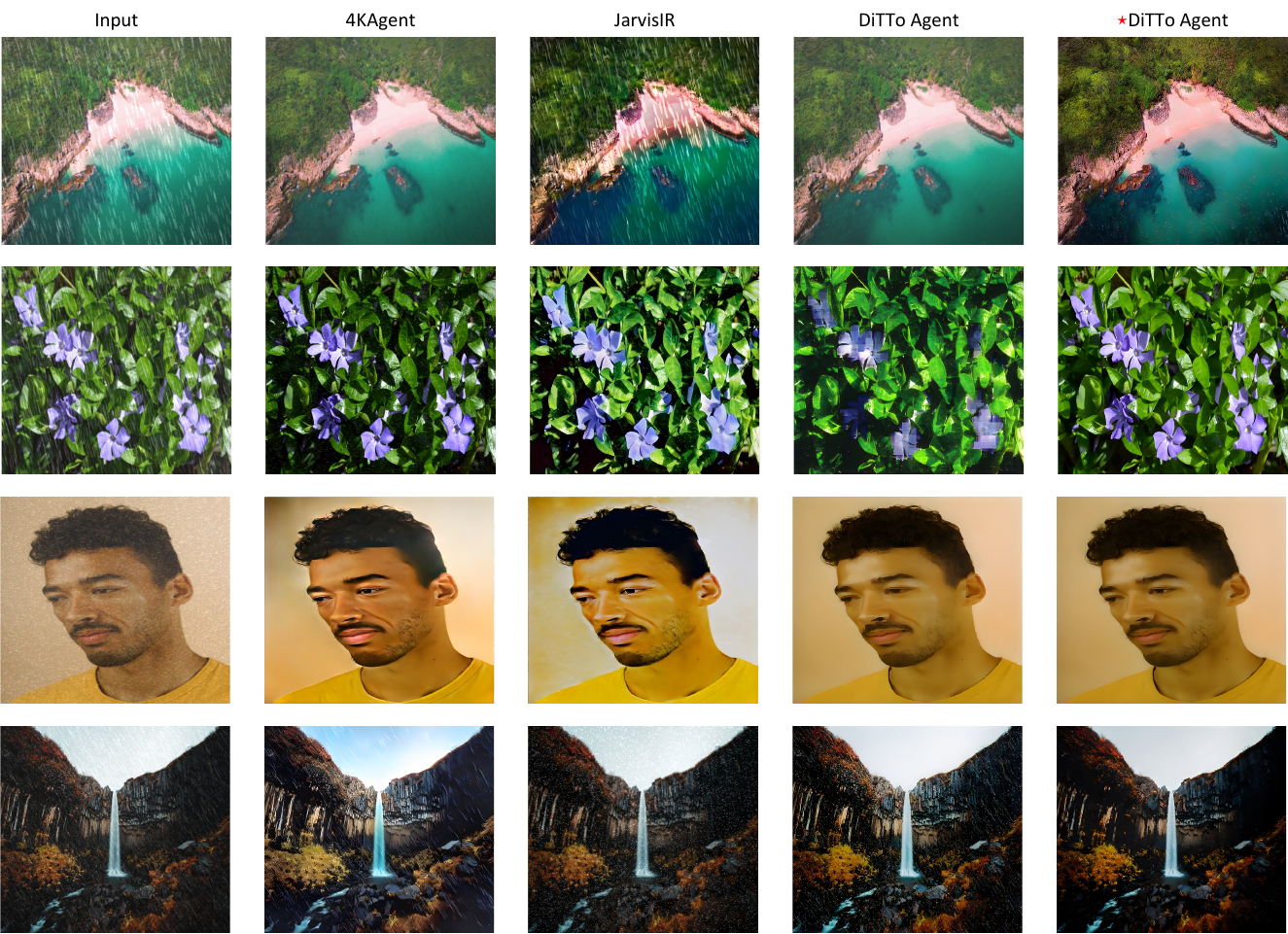}
    \caption{
    Additional qualitative comparisons on multi-degraded inputs with $j\!\in\!\{3,4,5\}$ concurrent degradations.}
    \label{fig:qualitative_high_j}
\end{figure}
\section{Extensibility Experiments}
\label{sec:extensibility}

This section reports the two extensibility scenarios that motivate the plug-and-play design of DiTTo: extending the restoration-expert pool $\{i^E_D\}$ for an existing degradation type, and extending the degradation universe $\mathbf{D}$ with a new degradation type.
In both scenarios, $\cup$S-IR, AiO-IQA, and $\mathcal{W}^{\text{SFT}}_{\text{DiTTo}}$ are reused, and only the efficient ORA stage is updated on a small expert-executed subset.

\subsection{Adding a New Restoration-Expert}
\label{sec:ext_expert}

We start from the $\star$DiTTo Agent configuration (Tab.~\ref{tab:expert_pool}) and add one further restoration-expert per degradation type to evaluate scalability of the ORA-only update.
We compare against a training-based agent baseline (JarvisIR) on the same held-out evaluation set, restricted to instances whose involved type set $\mathbf{D}^{\boldsymbol{\delta}}$ contains the affected degradation type so that the newly added restoration-expert is exercised.

\subsection{Adding a New Degradation Type}
\label{sec:ext_degradation}

We extend the degradation universe $\mathbf{D}$ with blur as an additional degradation type, motivated by the limitation discussion in Sec.~\ref{sec:limitations}.
Adding a new degradation type is strictly more demanding than adding a restoration-expert because it expands $\mathbf{D}$ itself, which in turn enlarges the candidate restoration-action set $\mathbf{A}^{\boldsymbol{\delta}}_{i^R}$ at every restoration index.
Concretely, we (i) augment the degradation synthesis pipeline (Sec.~\ref{sec:data_degsynth}) with a blur degradation-action $A^D_{\text{blur}}(\cdot)$, (ii) add a corresponding restoration-expert into the pool, (iii) regenerate a small expert-executed subset that includes blur in its involved type sets, and (iv) update $\mathcal{W}^{\text{ORA}}_{\text{DiTTo}}$ via the same ORA procedure as the main experiments.

\paragraph{Baseline protocol.}
For both scenarios, JarvisIR shares the SFT stage with DiTTo for parity, since SFT is not the locus of our contribution.
We run both JarvisIR and DiTTo under the same wall-clock budget on identical hardware ($2\times$B200).
Under this budget, DiTTo's ORA converges to completion, whereas JarvisIR's MRRHF alignment remains prohibitively slow due to repeated real-expert execution and multi-metric IQA scoring at every step.
We therefore report the restoration quality achieved by each method within the same practical adaptation budget.

\begin{table}[h]
\centering
\caption{Extensibility experiments. We compare DiTTo against a training-based baseline (JarvisIR) under two scenarios: (i) adding a new restoration-expert to an existing degradation type, and (ii) adding a new degradation type (blur) to $\mathbf{D}$. Both methods are evaluated under the same wall-clock adaptation budget on identical hardware ($2\times$B200), since fully running JarvisIR's MRRHF alignment under a changed expert pool is prohibitively slow. Higher is better for MUSIQ, MANIQA, CLIP-IQA; lower is better for NIQE.}
\label{tab:extensibility}
\small
\resizebox{\linewidth}{!}{%
\begin{tabular}{llcccc}
\toprule
Scenario & Method & MUSIQ\,$\uparrow$ & MANIQA\,$\uparrow$ & CLIP-IQA\,$\uparrow$ & NIQE\,$\downarrow$ \\
\midrule
\multirow{2}{*}{Add a new restoration-expert}
 & JarvisIR        & 59.91 & 0.480 & 0.642 & 7.63 \\
 & \textbf{DiTTo Agent} & \textbf{66.03} & \textbf{0.579} & \textbf{0.751} & \textbf{5.36} \\
\midrule
\multirow{2}{*}{Add a new degradation type (blur)}
 & JarvisIR        & 56.30 & 0.418 & 0.574 & 8.59 \\
 & \textbf{DiTTo Agent} & \textbf{62.93} & \textbf{0.530} & \textbf{0.697} & \textbf{5.84} \\
\bottomrule
\end{tabular}}
\end{table}

\paragraph{Discussion.}
Across both extensibility scenarios, DiTTo outperforms JarvisIR on every IQA metric.
The gap is consistent in direction (DiTTo $>$ JarvisIR) and substantial in magnitude on MUSIQ and CLIP-IQA, indicating that DiTTo's plug-and-play update produces visibly cleaner restored image-states under the same hardware budget.
The gap widens slightly in the harder scenario of adding a new degradation type, where JarvisIR drops on every metric while DiTTo retains most of its restoration quality.
This is consistent with our design: JarvisIR's online MRRHF alignment requires repeated real-expert execution and multi-metric IQA scoring at every step, so a fixed wall-clock budget covers fewer effective alignment updates as the candidate set $\mathbf{A}^{\boldsymbol{\delta}}_{i^R}$ enlarges.
DiTTo's ORA, by contrast, operates on pre-computed ORTD pairs and is unaffected by the per-step real-expert cost, so it converges within the same budget regardless of whether the extension targets the expert pool or the degradation universe.
The results validate the central claim of DiTTo: decoupling agent training from real-expert calls makes plug-and-play extension practical, both when the expert pool grows and when the degradation universe itself expands.



\section{Project Page}
\label{sec:demo}

We provide an anonymous project page that presents the DiTTo Agent
end-to-end inference pipeline in an interactive form. The page includes
drag-to-compare sliders between multi-degraded inputs and DiTTo-restored
outputs, a step-by-step visualization of the agent's reasoning
(degradation identification, order-aware restoration-action planning as
JSON-based tool calls, and sequential invocation of restoration-experts
with their intermediate restored image-states), qualitative comparisons
against prior agents, and short screen-recorded demo videos of the full
restoration loop on real multi-degraded images. All visualizations use the
same \scalebox{1.2}{\color{sh_red}{$\star$}}\textbf{DiTTo Agent} checkpoint
reported in Tab.~\ref{perception_table}. The project page is available at
\url{https://cmlab-korea.github.io/DiTTo/}.

\section{Limitations}
\label{sec:limitations}

DiTTo decouples agent training from the real restoration-expert pool
through $\cup$S-IR and AiO-IQA, but a residual
\emph{simulator-to-expert distribution gap} remains: ORA can only
correct the gap to the extent that
$\mathcal{D}_{\text{ORTD}}^{\text{Expert}}$ covers it.
Second, AiO-IQA mostly inherits its supervisory signal from
NR-IQA metrics, which are correlated with but not identical to
human perceptual quality, so atypical degradations (severe motion
blur, JPEG block artefacts beyond our universe $\mathbf{D}$) are not
directly handled.

\section{Broader Impacts}
\label{sec:broader}

DiTTo improves the quality of multi-degradation image restoration,
which has direct positive applications in autonomous driving
perception, mobile photography, and historical photo restoration.
Potential negative uses include enhancing surveillance imagery; we
do not release any new weights tied to identifiable persons, and the
training corpus consists only of public image datasets with
permissive licences.
The agent itself does not generate new content beyond restoring
inputs, which limits misuse for synthetic-media generation.

A specific privacy concern is the potential restoration of intentionally
obfuscated regions such as heavy mosaicking or blurring applied to faces
or license plates. We note, however, that such heavy obfuscation removes
most of the underlying signal, so any output produced from such inputs
is closer to model hallucination than to faithful recovery, and should
not be treated as identifying evidence. DiTTo is trained on the
degradation universe $\mathbf{D}$ defined in Sec.~\ref{sec:data_degsynth},
which does not include such severe privacy-protective obfuscation,
and we do not advocate using DiTTo Agent in any identification
or surveillance pipeline.



\end{document}